\documentclass[sigconf]{acmart}
 
\AtBeginDocument{%
  \providecommand\BibTeX{{%
    \normalfont B\kern-0.5em{\scshape i\kern-0.25em b}\kern-0.8em\TeX}}}

\copyrightyear{2020} 
\acmYear{2020} 
\setcopyright{acmlicensed}\acmConference[CIKM '20]{Proceedings of the 29th ACM International Conference on Information and Knowledge Management}{October 19--23, 2020}{Virtual Event, Ireland}
\acmBooktitle{Proceedings of the 29th ACM International Conference on Information and Knowledge Management (CIKM '20), October 19--23, 2020, Virtual Event, Ireland}
\acmPrice{15.00}
\acmDOI{10.1145/3340531.3411940}
\acmISBN{978-1-4503-6859-9/20/10}
\usepackage{color}
\usepackage{kotex}
\usepackage{comment}
\usepackage{amsmath,amssymb,amsfonts}
\usepackage{tabulary}
\usepackage{multirow}
\usepackage{pifont}
\usepackage{algorithmic}    
\usepackage{xspace}
\usepackage{xspace}

\newcommand{\toolname}{ST-GRAT\xspace}

\begin{document}

%%
%% The "title" command has an optional parameter,
%% allowing the author to define a "short title" to be used in page headers.
\title{\toolname: A Novel Spatio-temporal Graph Attention Network for Accurately Forecasting Dynamically Changing Road Speed}

%%
%% The "author" command and its associated commands are used to define
%% the authors and their affiliations.
%% Of note is the shared affiliation of the first two authors, and the
%% "authornote" and "authornotemark" commands
%% used to denote shared contribution to the research.

\settopmatter{authorsperrow=1}
\author{ $^1$Cheonbok Park$^\dagger$,  $^2$Chunggi Lee,  $^3$Hyojin Bahng,  $^3$Yunwon Tae,  $^4$Seungmin Jin,  $^2$Kihwan Kim, $^2$Sungahn Ko$^\ast$, and $^5$Jaegul Choo$^\ast$}
\affiliation{ $^1$ NAVER Corp, cbok.park@navercorp.com}
\affiliation{ $^2$ Ulsan National Institute of Science and Technology~(UNIST),~\{cglee, kh1875, sako\}@unist.ac.kr}
\affiliation{ $^3$ Korea University,~\{hjj522, tyj204\}@korea.ac.kr}
\affiliation{$^4$National Research University Higher School of Economics~(NRU-HSE),~sdzhin@hse.ru}
\affiliation{ $^5$ Korea Advanced Institute of Science and Technology~(KAIST),~jchoo@kaist.ac.kr}

\thanks{
$\ast$ Corresponding Authors\\
$\dagger$ Work done in korea University.
}

%\orcid{1234-5678-9012}
%%\author{G.K.M. Tobin}
%\authornotemark[1]
%\email{webmaster@marysville-ohio.com}

%%
%% By default, the full list of authors will be used in the page
%% headers. Often, this list is too long, and will overlap
%% other information printed in the page headers. This command allows
%% the author to define a more concise list
%% of authors' names for this purpose.
\renewcommand{\shortauthors}{Trovato and Tobin, et al.}
\fancyhead{}
%%
%% The abstract is a short summary of the work to be presented in the
%% article.
\settopmatter{printacmref=false, printfolios=false}
\begin{abstract}
Predicting road traffic speed is a challenging task due to different types of roads, abrupt speed change and spatial dependencies between roads; it requires the modeling of dynamically changing spatial dependencies among roads and temporal patterns over long input sequences. 
This paper proposes a novel spatio-temporal graph attention (\toolname) that effectively captures the spatio-temporal dynamics in road networks. The novel aspects of our approach mainly include spatial attention, temporal attention, and spatial sentinel vectors. 
The spatial attention takes the graph structure information (e.g., distance between roads) and dynamically adjusts spatial correlation based on road states.
The temporal attention is responsible for capturing traffic speed changes, and the sentinel vectors allow the model to retrieve new features from spatially correlated nodes or preserve existing features. 
The experimental results show that \toolname outperforms existing models, especially in difficult conditions where traffic speeds rapidly change (e.g., rush hours). 
We additionally provide a qualitative study to analyze \emph{when} and \emph{where} \toolname tended to make accurate predictions during rush-hour times.
\end{abstract}

%%
%% The code below is generated by the tool at http://dl.acm.org/ccs.cfm.
%% Please copy and paste the code instead of the example below.
%%


%%
%% Keywords. The author(s) should pick words that accurately describe
%% the work being presented. Separate the keywords with commas.

%% A "teaser" image appears between the author and affiliation
%% information and the body of the document, and typically spans the
%% page.
%%
%% This command processes the author and affiliation and title
%% information and builds the first part of the formatted document.
\maketitle
{\fontsize{8pt}{8pt} \selectfont
\textbf{ACM Reference Format:} \\
\noindent Cheonbok Park, Chunggi Lee, Hyojin Bahng, Yunwon Tae, Seungmin Jin, Kihwan Kim, Sungahn Ko, and Jaegul Choo. 2020. ST-GRAT: A Novel Spatio-temporal Graph Attention Network for Accurately Forecasting Dynamically Changing Road Speed. In \textit{Proceedings of the 29th ACM International Conference on Information and Knowledge Management (CIKM ’20), October 19–23, 2020, Virtual Event, Ireland.}  ACM, NY, NY, USA, 10 pages. https://doi.org/10.1145/3340531.3411940 }
\section{Introduction}
Predicting traffic speed is a challenging task, as a prediction method needs not only to find innate spatial-temporal dependencies among roads, but also needs to understand how these dependencies change over time and influence other traffic conditions.
For example, when a road is congested, there is a high chance that its neighboring roads are also congested.
Moreover, roads in residential areas tend to have different traffic patterns compared to those surrounding industrial complexes~\cite{Lee19}. 

Numerous deep learning models~\cite{Zhao17,Pan2018HyperSTNetHF} have been proposed for traffic speed prediction based on graph convolution neural networks (GCNNs) with recurrent neural networks (RNNs), outperforming conventional approaches~\cite{Vlahogianni14}. 
For example, a diffusion convolution recurrent neural network (DCRNN)~\cite{li2018dcrnn} combines diffusion convolution~\cite{atwood2016difconv} with an RNN and demonstrates improved prediction accuracy.
Graph WaveNet~\cite{Wu2019GraphWF} adapts diffusion convolution, incorporates a self-adaptive adjacency matrix, and uses dilated convolution for achieving state-of-the-art performance. 
However, the models assume \emph{fixed} spatial dependencies among roads, so that they compute spatial dependencies once and use the computed dependencies all the time without considering dynamically changing traffic conditions.

% However, the models assume fixed spatial dependencies among roads so that they can compute spatial dependencies once and use the computed dependencies from then on without considering dynamically changing traffic conditions.
%A recent study~\cite{li2018dcrnn} proposes a diffusion convolution recurrent neural network (DCRNN) that combines diffusion convolution~\cite{atwood2016difconv} with RNN and demonstrates improved prediction accuracy.
%Although the prior models show high performance, they also have weaknesses.
%First, existing GCNNs-based models (e.g., DCRNN~\cite{li2018dcrnn}, STGCN~\cite{yu2018spatio}, Graph WaveNet~\cite{Wu2019GraphWF}) assume \emph{fixed} spatial dependencies among roads. 
%Thus they compute spatial dependencies once and use the computed dependencies all the time without considering dynamically changing traffic conditions. 
To address this issue, recent models~\cite{Zhang2018gaan,zheng2019gman} utilize multi-head attention~\cite{Vas2017transformer} to model spatial dependencies. 
% However, these are only partial solutions as they do not consider the overall graph-structure information (e.g., distances, connections, and directions between nodes) that can play an important role in deciding which road should receive attention.
Despite efforts, they are only partial solutions, as they do not consider overall graph structure information (e.g., distances, connections and directions between nodes), which can play an important role in deciding which road to attend. 
In summary, any other approaches do not consider both of dynamically adjusting attention weights and the graph structure information.
%However, this work is a partial solution due the fact that does not consider overall graph structure information (e.g., distances, connections and directions between nodes), which can play an important role in deciding which road to attend. In summary, any other approaches do not consider both of dynamically adjusting attention weights and the graph structure information.

%Second, existing attention based spatial modelings does not consider traffic flow directions and overall graph structure information (e.g., distances, connections and directions between nodes), which can play an important role in deciding which road to attend.
%GaAN~\cite{Zhang2018gaan} and GMAN~\cite{zheng2019gman} apply different spatial correlation results across roads by utilizing an attention mechanism. 
%However, existing attention based spatial modelings does not consider traffic flow directions and overall graph structure information (e.g., distances, connections and directions between nodes), which can play an important role in deciding which road to attend. 
%\cb{For examples, GaAN~\cite{Zhang2018gaan} just utilizes connectivity information when computing attention in the spatial modeling.  GMAN~\cite{zheng2019gman} also calculates spatial dependencies regardless of connectivity and direction between nodes. }
% \cb{}

Many models employ recurrent neural networks (RNNs) for temporal modeling (e.g., DCRNN~\cite{li2018dcrnn}, GaAN~\cite{Zhang2018gaan}).
However, RNNs have a limitation in that they cannot directly access past features in long input sequences, which implies ineffectiveness in modeling temporal dependencies~\cite{Wu2019GraphWF}.
%But RNNs have a limitation in directly accessing past features in long input sequences, so they cannot effectively model the temporal dependencies~\cite{Wu2019GraphWF}.
%Second, there are models that use recurrent neural networks (RNNs) for temporal modeling (e.g., DCRNN~\cite{li2018dcrnn} GaAN~\cite{Zhang2018gaan}).
%However, RNNs cannot directly access past features in long input sequences, which implies a limitation in effectively encoding temporal dynamics}~\cite{Wu2019GraphWF}.
Attention-based models can be an alternative to resolve the issue in the RNN-based temporal modeling, directly accessing past information in the long input sequences. 
However, existing attention models do not consider dynamic temporal dependencies among roads. 
For example, there are cases where road speed can be best predicted by attending the target roads. 
However, existing models with attention do not consider these cases, so that they always retrieve new information from neighbor nodes, even in unnecessary cases.
% However, existing models with attention do not consider these cases, so they always retrieve new information from the neighbor nodes, even when it is unnecessary.
%new information from other roads or focus on the existing encoded features.

%Third, existing attention techniques~\cite{Vas2017transformer,Zhang2018gaan,zheng2019gman} force a model to always retrieve new information from neighbor nodes, even in unnecessary cases. It makes the model to disturb predictions by extracting unrelated features.

In this work, we propose a novel spatio-temporal graph attention network (\toolname) for predicting traffic speed that addresses aforementioned weaknesses. 
First, we design a spatial attention module to model the spatial dependencies by capturing both road speed changes and graph structure information, based on our proposed diffusion prior, directed heads, and distance embedding. 
%Inspired by GCNNs, we distill the knowledge of diffusion process of traffic networks in capturing spatial dependencies.  
Second, we encode temporal dependencies by using attention to directly access distant relevant features of input sequences without any restriction and to effectively capture sudden fluctuating temporal dynamics.
%unlike the aforementioned models that use RNNs for temporal modeling, \toolname can directly access distant relevant features of input sequences without any restriction by using temporal attention. It effectively encodes temporal dynamics and reflects sudden fluctuating temporal dynamics.
Third, to avoid attending unrelated roads that are not helpful for prediction, we newly design `spatial sentinel' key and value vectors, motivated by the sentinel mixture model~\cite{Merity2016PointerSM,Lu2016KnowingWT}.
%Third, to remove attend unrelated nodes, we newly design `spatial sentinel' key and value vectors to the spatial attention, which is motivated by the sentinel mixture model~\cite{Merity2016PointerSM,Lu2016KnowingWT}.
Guided by the sentinel vectors, \toolname dynamically decides to use new information of other roads or focus on existing encoded features. 
%The sentinel vectors allow \toolname to avoid unnecessary attention and focus on existing encoded features instead.
%new information from other roads or focus on the existing encoded features.
The experimental results indicate that \toolname achieves state-of-the-art performance, especially in short-term prediction. % and abruptly traffic speed change condition.  
We also confirm that \toolname is better than existing models at predicting traffic speeds in situations where the road speeds are abruptly changing (e.g., rush hours). % and short-term prediction.
Moreover, compared to existing methods, our model shows an interptetable ability by using the self-attention mechanism with the sentinel vectors. In the qualitative study, we conduct an interpretation of our trained model by visualizing when and where the model directed its attention based on different traffic conditions.
Lastly, we present in-depth alalyses on how the newly designed components dynamically capture spatio-temporal dependencies. 
%In the qualitative study, we conducted an interpretation of our trained model by visualizing when and where the model directed its attention based on different traffic conditions.

The contributions of this work include: 
\begin{itemize}
\item \toolname, which consists of entire self-attention mechanisms to dynamically captures both spatial and temporal dependencies of input sequences over time. 
\item A newly proposed self-attention module with the sentinel vectors that help the model decide to focus on existing encoded features, instead of unnecessary attending other roads, 
\item A spatial module that uses diffusion prior and directed heads to effectively encode graph structure, 
%Our spatial information exploits the graph structure information by using diffusion prior and directed heads.
\item Quantitative experiments and comparisons with state-of-the art models on the two real-world datasets, including the different time ranges and abruptly changing time ranges (e.g., rush hours), and 
\item In-depth analysis and interpretation on how \toolname works in varying traffic conditions
%The qualitative study, which describes why newly designed features of \toolname are important in varying traffic conditions
\end{itemize}

\section{Related Work}
 
In this section, we review previous approaches regarding traffic prediction and attention models.

%\subsection{Approaches for Traffic Forecasting}

\subsection{Approaches for Traffic Forecasting} 
Deep learning models for traffic prediction usually leverage spatial and temporal dependencies of the road traffic.
The graph convolution neural network (GCNN)~\cite{kipf2016semi} have been popular for spatial relationship modeling. 
Given a road network, it aggregates adjacent node information into features based on convolution coefficients. 
These coefficients are computed by spatial information (e.g., the distance between nodes).
RNNs and their variants are combined with the encoded spatial relationship to model temporal dependencies (e.g., speed sequences)~\cite{yu2018spatio}.

As modeling spatial correlation is a key factor for improving prediction performance, researchers have proposed new approaches for effective spatial correlation modeling.
For example, DCRNN ~\cite{li2018dcrnn} combines diffusion convolution~\cite{atwood2016difconv} and recurrent neural networks to model spatial and temporal dependencies. 
Graph WaveNet~\cite{Wu2019GraphWF} also adapts diffusion convolution in spatial modeling, but it is different from DCRNN, as it 1) considers both connected and unconnected nodes in the modeling process, and 2) uses dilated convolution~\cite{vanwavenet} to learn long sequences of data.

Nonetheless, existing approaches use constant coefficients, which are computed once and applied to all traffic conditions. However, the fixed coefficients may result in inaccuracies when spatial correlation is variable (e.g., abrupt speed changes).
Compared to existing models, \toolname improves accuracy by dynamically adjusting the coefficients of neighboring nodes based on their present states and more spatial information (e.g., distance, node connectivity, flow direction).

\subsection{Attention Models}

Attention-based neural networks are widely used for sequence-to-sequence modeling, such as machine translation and natural language processing (NLP) tasks~\cite{bahdnamu2014seq2attn,Xu2015ShowAA,lin2017structured,seo2016bidirectional}. 
Vaswani et al. propose a novel self-attention network called Transformer~\cite{Vas2017transformer}, which is able to dynamically capture diverse syntactic and semantic features of the given context by using multi-head self-attention heads. 
The self-attention mechanism has additional advantages compared to conventional long-short-term memory (LSTM)~\cite{hochreiter1997lstm} in that its process can be easily paralleled, and it directly attends to related input items regardless of the coverage of the receptive field. 
Due to these advantages, Transformer has contributed to many other NLP tasks for improving accuracy~\cite{radford2019gpt2,jacob2018bert}.
Another study~\cite{Veli2018gat} used the self-attention network for graph data, demonstrating that the attention networks outperform the GCNN model. 
Zhang et al.~\cite{Zhang2018gaan} propose a graph attention network, replacing the diffusion convolution operation in DCRNN~\cite{li2018dcrnn} with the gating attention. These models show that the graph attention model does not lag behind the GCNN-based model in the spatio-temporal task.

While previous models can be used for replacing GCNN-based spatial modeling, 
%they all have a drawback--they do not consider the information embedded in the graph structure in deciding when and where to attend, such as distances, connectivity and flow directions between nodes in their spatial dependency modeling processes.
they all have a drawback; they do not consider the information embedded in the graph structure (such as distances, connectivity, and flow directions between nodes) in their spatial dependency modeling processes in deciding when and where to attend.
%This means, when they decide which node to attend, they do not consider proximity between nodes, which is an important factor to capture spatial relationship on adjacent nodes. 
Compared to previous models, \toolname has a novel spatial attention mechanism that can consider all of the mentioned graph structure information.

\section{Proposed Approach}
In this section, we define the traffic forecasting problem and describe our spatio-temporal graph attention network.
\begin{figure}
    \centering
    \includegraphics[width=.95\columnwidth]{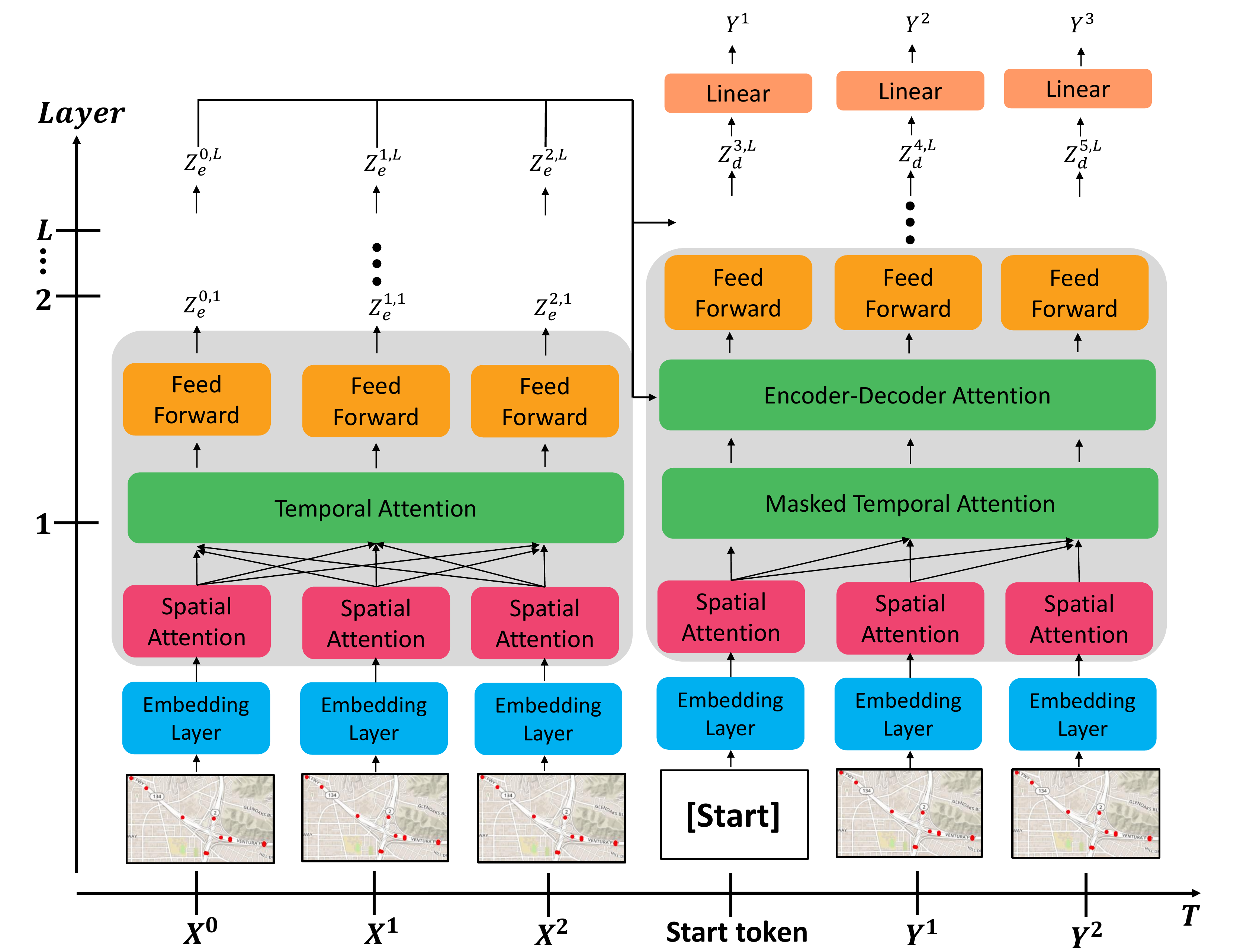}
    \caption{Overall architecture of \toolname. The $x$ and $y$ axes indicate the time and the number of the layers, respectively. The left-half block is the $L$-stacked encoder while the right-half block is the $L$ stacked decoder. We use a special token, `[Start]', to represent the starting point in a decoding stage.}
    \label{fig:arch}
\end{figure}
\subsection{Problem Setting for Traffic Speed Forecasting} We aim to predict the future traffic speed at each sensor location. 
We define the input graph as $\mathcal{G} = (\mathcal{V},\mathcal{E},\mathcal{A})$, where $\mathcal{V}$ is the set of all the different sensor nodes, ($|\mathcal{V}|=N$), $\mathcal{E}$ is the set of edges, and $\mathcal{A} \in \mathbb{R}^{N \times N}$ is a weighted adjacency matrix.
The matrix $\mathcal{A}$ includes two types of proximity information in the road network: connectivity and edge weights.
%The matrix $\mathcal{A}$ includes three types of information: connectivity, edge weights, and proximity. 
Connectivity indicates whether two nodes are directly connected or not. Edge weights are comprised of the distance and direction of the edges between two connected nodes. This proximity information refers to the overall structure on a given graph including connectivity, edge directions, and distances of the entire nodes. 

We denote $X^{\left(t\right)} \in \mathbb{R}^{N \times 2}$ as the input feature matrix at time $t$, where $N$ is the number of nodes and $2$ is the number of features (the velocity and the timestamp). 
% Following the general problem setting in most of previous traffic forecasting frameworks , which is  follows most of the previous the  
Following the conventional traffic forecasting problem definition, our problem is to learn a mapping function $f$ that predicts the speed of the next $T$ time steps ($Y=[X^{\left(t+1\right)}_{:,0},\cdots,X^{\left(t+T\right)}_{:,0}]$), given the previous $T$ input speeds in a sequence ($X=[X^{\left(t-T+1\right)},\cdots,X^{\left(t\right)}]$), and graph $\mathcal{G}$, i.e.,  $Y=f(X,\mathcal{G}) $.
To solve this sequence-to-sequence learning problem, we utilize an encoder-decoder architecture, as shown in Fig.~\ref{fig:arch} described in the following sections.
%which is described in the following sections.

\subsection{Encoder Architecture} 
Given a sequence of observations, $X$, the encoder consists of spatial attention and temporal attention for predicting the future sequence $Y$. 
As shown in Fig.~\ref{fig:arch}, a single encoder layer consists of three sequential sub-layers: the spatial attention layer, the temporal attention layer, and the point-wise feed-forward neural networks. 
The spatial attention layer attends to neighbor nodes spatially correlated to the center node at each time-step, while the temporal attention layer attends to individual nodes and focuses on different time steps of a given input sequence. The position-wise feed-forward networks create high-level features that integrate information from the two attention layers. These layers consist of two sequential fully connected networks with GELU~\cite{hendrycks2016gelu} activation function.

The encoder has a skip-connection to bypass the sub-layer, and we employ layer normalization~\cite{Ba2016LayerN} and dropout after each sub-layer to improve the generalization performance. 
The overall encoder architecture is a stack of an embedding layer and four ($L=4$) identical encoder layers. 
The encoder transforms the spatial and temporal dependencies of an input signal into a hidden representation vector, which is used later for attention layers in the decoder.

\subsection{Embedding Layer}
Unlike GCNN-based models, attention-based GNNs~\cite{Veli2018gat,Zhang2018gaan} mainly utilize connectivity between nodes.
However, conventional models do not consider proximity information in their modeling process.
To incorporate the proximity information, the embedding layer in \toolname takes a pre-trained node-embedding vector generated by LINE~\cite{Tang2015LINE}.
The node-embedding features are used to compute spatial attention, which will be further discussed in the following section.

The embedding layer also performs positional embedding to acquire the order of input sequences.
Unlike previous methods that use a recurrent or convolutional layer for sequence modeling, we follow the positional encoding scheme of the Transformer~\cite{Vas2017transformer}.% Positional encoding does not require extra training parameters. 
 We apply residual skip connections to prevent the vanishing effect of embedded features that can occur as the number of encoder or decoder layers increases. 
We concatenate each node embedding result with the node features and then project the concatenated features onto $d_{model}$. 
Lastly, we add the positional encoding vector to each time step.

\begin{figure}
    \centering
    \includegraphics[width=1\columnwidth]{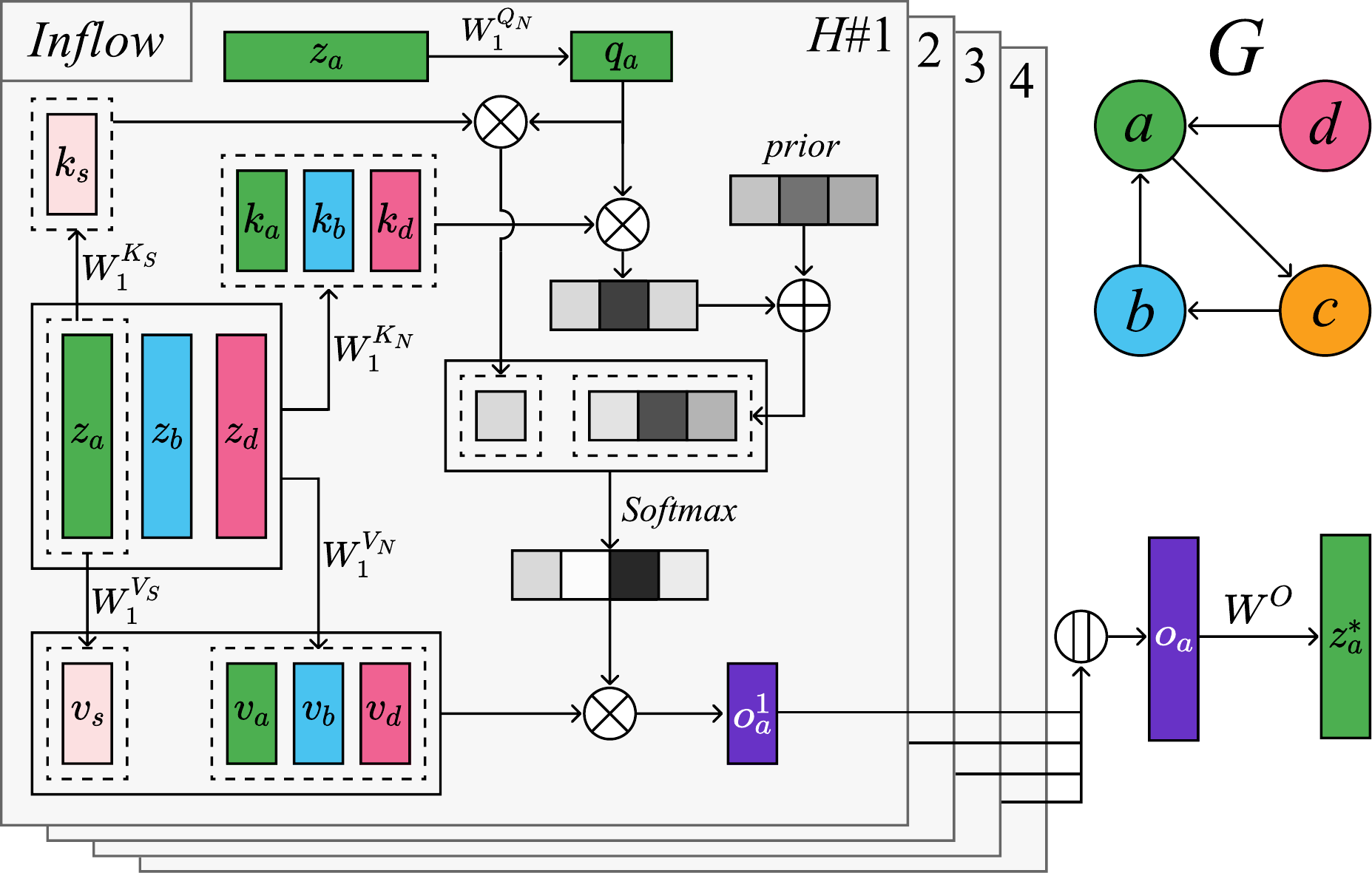}
    \caption{The proposed spatial attention mechanism. In this example, the inflow spatial attention takes query, key, and value vectors, $q$, $k$, and $v$, respectively. $k_{s}$ and $v_{s}$ indicate a sentinel key and value vector. $z^{*}$ represents the output of multi-head attention, and $H\#$ is an indicator of the heads. Lastly,  $\oplus$, $\otimes$, and $\parallel$ indicate the element-wise sum, the matrix multiplication, and the concatenation operation, respectively.}
    \label{fig:node_attn}
\end{figure}

\subsection{Spatial Attention} Fig.~\ref{fig:node_attn} shows the proposed spatial attention, which consists of multiple inflow attention heads~(odd head indices) and outflow attention heads~(even head indices).  
Previous attention-based GNNs~\cite{Zhang2018gaan,Veli2018gat} define spatial correlation in an undirected manner. They calculate attention with respect to all neighbor nodes without considering their direction in a road network. In contrast, our model differentiates neighbor nodes by  directions of attention heads.
%In contrast, our model differentiates neighbor nodes by direction (i.e., in-coming and out-going). 
Specifically, we divide the attention heads--i.e., odd indices are responsible for inflow nodes, even indices are responsible for outflow nodes—which allows the model to attend to different information for each of inflow and outflow traffic. Fig.~\ref{fig:node_attn} shows the inflow attention head example, which attends only inflow nodes and its node.
% On the other hand, our model specifically divides the neighbor nodes into in-coming and out-going traffic with respect to each node by head index. By dividing the attention heads, the model is able to attend to different information for each of inflow and outflow traffic.

%Each attention head aggregates the information of a current node and its neighbor nodes given states of each node. 
We denote the encoder hidden states as $\mathcal{Z} = [z_{i},\cdots,z_{N}]$,  where $z_{i} \in \mathcal{R}^{ d_{model}}$ is the hidden state of the $i$-th node. We denote the set of the $i$-th node and its neighbor nodes as $\mathcal{N}_{i}$. We define the dimensions of the query, key, and value vectors as $d_{q}=d_{k}=d_{v} = d_{model}/H$, respectively, where $H$ is the number of attention head in multi-head self-attention.
%Each attention head aggregate the information of a current $i$-th node and its neighbor nodes given states of each node. 

To extract diverse high-level features from multiple attention heads, we project the current node onto a query space and $\mathcal{N}_{i}$ onto the key and value spaces.  
The output of each attention head is defined as a weighted sum of value vectors, where the weight of each value is computed from a learned similarity function of the corresponding key with the query. %In our spatial attention, the weights is an attention of the query, and is considered as a spatial correlation of the query nodes in the graph.
However, existing self-attention methods have the constraint that the sum of the weights has to be one. Hence, the query node has to attend to key-value pairs of $\mathcal{N}_{i}$, even in situations where any spatial correlation does not exist among them.
%This characteristic leads to unnecessary attention weights when non spatial correlation in the key-value pairs of $\mathcal{N}_{i}$. 
%However, this characteristics leads to attention is not able to attend noting and has unnecessary attention when no relevant features in the key-value pairs. 
%To alleviate this issue, we add `` spatial sentinel'' key and value vectors as an additional key-value pair, which is an alternative feature when the model decides to do not extract new information from the key-value pair of $\mathcal{N}_{i}$. If the query node is unnecessary to attend to the key-value pairs of $\mathcal{N}_{i}$, the query node only attends the sentinel value vector. In each attention, we the sentinel value and key vector are linear transformation vectors of a query node.

To prevent such unnecessary attention, we add spatial sentinel key and value vectors, which are linear transformations of a query node. For instance, if a query node does not require any information from the key-value pairs of $\mathcal{N}_{i}$, it will attend to the sentinel key and value vectors (i.e., stick to its existing information rather than always attending to the given key and the value information).

Thus, the output feature of the $i$-th node in the $h$-th attention head, $o_{i}^{h} \in \mathcal{R}^{d_v}$, is the weighted sum of the spatial sentinel value vector and the value vectors of $\mathcal{N}_{i}$: 
\begin{comment}

\begin{equation}
\begin{array}{r@{}l}
o_{i}^{h} &= \sum_{j=\mathcal{N}_{i}}\alpha_{i,j}(z_{i}W_{i}^{V_s})+ \sum_{j=\mathcal{N}_{i}}\alpha_{i,j}\left(z_jW_{i}^{V_{s}}\right), \\
%\end{align*}
\end{array}
\end{equation}
\end{comment}

\begin{align*}
o_{i}^{h} &= \left(1-\sum_{j=\mathcal{N}_{i}}\alpha_{i,j}\right)*\left(z_{i}W_{h}^{V_s} \right) + \sum_{j=\mathcal{N}_{i}}\alpha_{i,j}\left(z_jW_{h}^{V_{n}}\right),     
\end{align*}
where  $W_{h}^{V_s} \in \mathcal{R}^{d_{model} \times d_{v}}$ and $W_{h}^{V_n} \in \mathcal{R}^{d_{model} \times d_{v}} $ indicate the linear transformation matrices of the sentinel value vector and the value vector of spatial attention, respectively. The attention coefficient, $\alpha_{i,j}$, is computed as %a softmax over the energy logits and the sentinel energy logit $e_{i,s}$,
\begin{equation}\label{eqn:node-2}
\begin{array}{r@{}l}
\alpha_{i,j} = \frac{\exp\left(e_{i,j}\right)}{e_{i,s}+\sum_{j=\mathcal{N}_{i}}\exp\left(e_{i,j}\right)},
\end{array}
\end{equation}
where $e_{i,j}$ indicates the energy logits, and $e_{i,s}$ represents the sentinel energy logit. 
%similarity score between the query and key vectors. 

Energy logits are computed by using a scaled dot-product of the query vector of the $i$-th node and the key vector of the $j$-th node, i.e., 
\begin{equation}\label{eqn:node-3}
\begin{array}{r@{}l}
e_{i,j} = \frac{\left(z_{i}W_{h}^{Q_{N}}\right)\left(z_{j}W_{h}^{K_{N}}\right)^T}{\sqrt{d_{k}}} + P_{h}(\mathcal{A}), 
\end{array}
\end{equation}
where parameter matrices $W_{h}^{V_s} \in \mathcal{R}^{d_{model} \times d_{q}}$ and $W_{h}^{V_n} \in \mathcal{R}^{d_{model} \times d_{k}} $ are the linear transformation matrices of the query vector and the key vector. %We additionally add a prior knowledge $P(\mathcal{A}_{h})$ when computing the energy logit of $\mathcal{N}_{i}$ in the $h$-th attention head. 
%We find that in spatial modeling, the nearest-neighbor nodes of the particular node are more influential than distantly located nodes.
%Diffusion process is the effective basis information of the spatial modeling
%Considering this property, we define an additional prior knowledge $P(\mathcal{A})$, called \textit{diffusion prior}, based on a diffusion process in the graph. The diffusion prior indicates whether the attention head is an out-going attention or an in-coming attention, i.e., 
Moreover, to explicitly provide edge information, we include additional prior knowledge $P_{h}(\mathcal{A})$, called \textit{diffusion prior}, based on a diffusion process in a graph. The diffusion prior indicates whether the attention head is an inflow attention or an outflow attention, defined as 
%Considering this property, we define an additional prior knowledge $P(\mathcal{A})$, called \textit{diffusion prior}, based on a diffusion process in the graph. The diffusion prior indicates whether the attention head is an out-going attention or an in-coming attention, i.e., 
\begin{equation}\label{eqn:node-4}
\begin{array}{r@{}l}
P_{2m+1}(\mathcal{A}) = \sum_{k=0}^{K}\beta^{k}_h*(D_I^{-1}\mathcal{A}^T)^k \\
P_{2m}(\mathcal{A}) = \sum_{k=0}^{K}\beta^{k}_h*(D_O^{-1}\mathcal{A})^k, 
\end{array}
\end{equation}
where $K$ is the number of truncation steps of the diffusion process. $D_{I}$ and $D_{O}$ are the in-coming diagonal matrix and out-going diagonal matrix respectively. $\left(D_{O}^{-1}\mathcal{A}\right)$ and $\left(D_{I}^{-1}\mathcal{A}\right)$ denote the out-going and the in-coming state transition matrices. %Thus, our prior is defined as the weighted sum of truncated state transitions. 
$\beta^{k}_{h}$ is the weight of the diffusion process at step $k$ in the $h$-th attention head, which is a learnable parameter at each layer of the attention head. 
%It helps refine the attention weights by using the prior. 

Calculating the sentinel energy logit is similar to other energy logits, but it excludes prior knowledge and uses a sentinel key vector instead. Hence, it is defined as follows:
\begin{equation}\label{eqn:node-5}
\begin{array}{r@{}l}
e_{i,s} =\frac{\left(z_{i}W_{h}^{Q_{N}}\right)\left(z_{i}W_{h}^{K_{s}}\right)^T}{\sqrt{d_{k}}},  \\
\end{array}
\end{equation}
% 하나의 attetnion을 사용하는 대신에 여러 attention을 사용함으로써 
where $W_{h}^{K_{s}} \in \mathcal{R}^{d_{model} \times d_{k}}$ is a linear transformation matrix of the sentinel key vector. For example, If $e_{i,s}$ is higher than $\sum e_{i,j}$, the model will assign less attention to the nodes in $\mathcal{N}_{i}$ nodes. 

After computing the output features $o_{i}^{h}$ on each attention head, they are concatenated and projected as
\begin{equation}\label{eq.node_output}
z_{i}^{*} = \mathrm{Concat}(o_{i}^{1},...,o_{i}^{H})W^{O_{N}},
\end{equation}
where $W^{O} \in \mathcal{R}^{d_{model} \times d_{model}}$ is the projection layer. The projection layer helps the model to combine various aspects of spatial-correlation features and the outputs of the inflow and outflow attention heads.

\begin{table*}[t]\caption{Prediction accuracy on METR-LA }
\centering
\resizebox{\textwidth}{!}{\begin{tabular}{l|l|l|lllllllll}
\hline
                   & T                        & Metric & GCRNN  & DCRNN   & GaAN   & STGCN   & Graph WaveNet  & HyperST & GMAN& \toolname                   \\ 
\hline
\multirow{12}{*}{\rotatebox[origin=c]{90}{METR-LA}} & \multirow{3}{*}{15 min}  & MAE        & 2.80   & 2.73    & 2.71   & 2.88    & 2.69 & 2.71   &                                           2.81 & \textbf{2.60}      \\
                   &                          & RMSE   & 5.51   & 5.27    & 5.24   & 5.74    & 5.15              & 5.23                                                    & 5.55& \textbf{5.07}      \\
                   &                          & MAPE    & 7.5\%  & 7.12\%  & 6.99\% & 7.62\%  & 6.90\%            & -                                                       & 7.43\% & \textbf{6.61}\%    \\ 
\cline{2-11}
                   & \multirow{3}{*}{30 min}  & MAE    & 3.24   & 3.13    & 3.12   & 3.47    & 3.07              & 3.12                                                    &3.12 & \textbf{3.01}      \\
                   &                          & RMSE    & 6.74   & 6.40    & 6.36   & 7.24    & 6.26              & 6.38                                                 &6.46    & \textbf{6.21}      \\
                   &                          & MAPE   & 9.0\%  & 8.65\%  & 8.56\% & 9.57\%  & 8.37\%            & -                                                      &8.35\%  & \textbf{8.15} \%   \\ 
\cline{2-11}
                   & \multirow{3}{*}{1 hour}  & MAE    & 3.81   & 3.58    & 3.64   & 4.59    & 3.53              & 3.58                                                   & \textbf{3.46} & 3.49      \\
                   &                          & RMSE   & 8.16   & 7.60    & 7.65   & 9.40    & \textbf{7.37}     & 7.56                                                &   \textbf{7.37}  & 7.42               \\
                   &                          & MAPE   & 10.9\% & 10.43\% & 10.62\% & 12.70\% & \textbf{10.01\%}  & -                                                 &10.06\%       & \textbf{10.01\%}   \\ 
\cline{2-11}
                   & \multirow{3}{*}{Average} & MAE    & 3.28   & 3.14    & 3.16   & 3.64    & 3.09              & 3.13                                         & 3.13  & \textbf{3.03}               \\
                   &                          & RMSE   & 6.80   & 6.42    & 6.41    & 7.46    & 6.26              & 
                   6.39 & 6.46 &\textbf{6.23}               \\
                   &                          & MAPE   & 9.13\% & 8.73\%  & 8.72\% & 9.96\%  & 8.42\%            & -                                             & 8.61\% & \textbf{8.25\%}             \\ 
\hline

\end{tabular}}
\label{table.metr}
%\vspace{-0.5cm}
\end{table*}

\begin{table}[t]\caption{Summary of experiment results on PEMS-Bay datasets.}
\centering
\resizebox{\columnwidth}{!}{\begin{tabular}{l|l|l|llllllll}
\hline
                   & T                        & Metric & DCRNN     & STGCN   & Graph WaveNet  &  GMAN& \toolname                   \\ 
\hline
\multirow{12}{*}{\rotatebox[origin=c]{90}{PEMS-Bay}} & \multirow{3}{*}{15 min}  
												  & MAE      & 1.38      & 1.36    & 1.30                                                                    & 1.36&  \textbf{1.29}                  \\
                   &                           & RMSE           & 2.95         & 2.96    & 2.74                                                                    &2.93 & \textbf{2.71}               \\
                   &                          & MAPE         & 2.9\%      & 2.9\%   & 2.73\%                                                               &2.88\%  & \textbf{2.67\%}                 \\ 
\cline{2-8}
                   & \multirow{3}{*}{30 min}  
                   								& MAE         & 1.74       & 1.81    & 1.63                                                                &1.64   & \textbf{1.61}                    \\
                   &                          & RMSE         & 3.97        & 4.27    & 3.70                                                                &3.78   & \textbf{3.69}                    \\
                   &                          & MAPE        & 3.9\%       & 4.17\%  & 3.67\%                                                            & 3.71\%   & \textbf{3.63\%}                    \\ 
\cline{2-8}
                   & \multirow{3}{*}{1 hour}  
                   								& MAE         & 2.07          & 2.49    &1.95                                                                  &\textbf{1.90}   &   1.95                \\
                   &                          & RMSE       & 4.74         & 5.69    & 4.52                                                                    &\textbf{4.40} & 4.54                   \\
                   &                          & MAPE         & 4.9\%       & 5.79\%  & 4.63\%                                                        &\textbf{4.45\%}         & 4.64\%                 \\ 
\cline{2-8}
                   & \multirow{3}{*}{Average} 
                   								& MAE           & 1.73        & 1.88    & 1.63&                      1.63                                         &\textbf{1.62}                  \\
                   &                          & RMSE       & 3.88        & 4.30   &\textbf{3.65}                                                   & 3.70&     \textbf{3.65}                  \\
                   &                          & MAPE        & 3.9\%      & 4.28\%  & 3.67\%     & 3.68\% &\textbf{3.65\%}     \\

\hline
\end{tabular}}
\label{table.pems}
\end{table}

\subsection{Temporal Attention} There are two major differences between temporal and spatial attention: 1) temporal attention does not use the sentinel vectors and the diffusion prior, and 2) temporal attention attends to important time steps of each node, while spatial attention attends to important nodes at each time step.
However, temporal attention is similar to spatial attention in that it uses multi-head attention to capture the diverse representation in the query, key, and value spaces. We utilize the multi-head attention mechanism, which is proposed in Transformer~\cite{Vas2017transformer}.

\begin{comment}

The temporal attention is computed by concatenating the output matrix of each attention head and projecting it by $W^{O_{T}} \in \mathcal{R}^{d_{model} \times d_{model}}$:
\begin{align*}
MultiHead &= \mathrm{Concat}(head_{1},...,head_{H})W^{O_{T}}.
\end{align*}
At each temporal attention head, $ \mathcal{Z}_{i} = [z_{i}^{(t-T+1)},\cdots,z_{i}^{t}]$ is projected onto the query, key, and value spaces, 
\begin{align*} 
head_{i} &= \mathrm{Attention}(QW_{i}^{Q_{T}},KW_{i}^{K_{T}},VW_{i}^{V_{T}}),
\end{align*}
where $\mathcal{H}_{i}$ is an input sequence of hidden state features of the $i$-th node, and $W_{i}^{Q_{T}}\in\mathcal{R}^{d_{model} \times d_{q}}$,$W_{i}^{K_{T}}\in\mathcal{R}^{d_{model} \times d_{k} }$ and  $W_{i}^{V_{T}} \in \mathcal{R}^{d_{model} \times d_{v}}$ indicates the linear transformation matrix of the query, key and value respectively. 
After projecting each time-step feature onto sub-spaces, we apply the scaled dot-product attention to obtain the attention weight at each time step.
We calculate the output matrix of the weighted sum of value vectors which is defined as follows:
\begin{align*} 
Attention(Q,K,V) &= \mathrm{Softmax}\left(\frac{QK^T}{\sqrt{d_{k}}} \right)V.
\end{align*}
\end{comment}

Note that the temporal attention layer can directly attend to features across time steps without any restriction in accessing information in the input sequence, which is different from previous approaches~\cite{li2018dcrnn,Zhang2018gaan} that cannot access features at distant time steps. 

\subsection{Decoder Architecture} 
The overall structure of the decoder is similar to that of the encoder. 
The decoder consists of the embedding layer and four other sub-layers: the spatial attention layer, two temporal attention layers, and the feed-forward neural networks. After each sub-layer, layer normalization is applied. 
One difference between the encoder and decoder is that the decoder contains two different temporal attention layers--the masked attention layer and the encoder-decoder (E-D) attention layer. 
The masked attention layer masks future time step inputs to restrict attention to present and past information. 
The encoder-decoder attention layer extracts features by using the encoded vectors from both the encoder and the masked attention layer. 
In this layer, the encoder-decoder attention from the encoder is used as both the key and the value, and query features of each node are passed along with the masked self-attention. 
Finally, the linear layer predicts the future sequence.

\begin{comment}

\subsection{Scheduled Sampling}

In the training phase, the model predicts a future time-step $Y^{(t)}$ at $t$ given a \textit{ground-truth} sequence $Y^{(:t-1)}$ as done in traditional sequence-to-sequence training approaches. On the other hand, in the inference phase, the model is given a \textit{predicted} output sequence of previous time-step values to predict the next time-step value. This significant input difference between two phases brings the amplified error from mistakes in intermediate time-steps. To alleviate this issue, we utilize curriculum learning~\cite{Bengio15scheduled}, which gradually decreases the ratio of the input ground-truth value $\epsilon_{i}$ in the training decoder at $i$-th iteration as 
\begin{equation}
\epsilon_i = \frac{\kappa}{\kappa+\exp^\frac{i}{\kappa}}
\end{equation}
,where $\kappa$ is the convergence speed parameter. This method makes the model robust against previous wrong predictions mistakes during the inference time, and reduces the disparity between the training and inference phase. In our experiments , we set $\kappa$ as 10000.

\end{comment}

\section{Evaluation}
We present our experimental results on two real-world large-scale datasets--METR-LA and PEMS-BAY released by ~\cite{li2018dcrnn}.

%The METR-LA and PEMS-BAY dataset are four months of 207 sensors' and six month of 325 sensors' speed data on the highways of Los Angeles County and in Bay area, respectively. 
The METRLA and PEMS-BAY datasets contain speed data for the period of four months from 207 sensors and six months from 325 sensors, gathered on the highways of Los Angeles County and in the Bay area, respectively.
We pre-process the data to have a five-minute interval speed and timestamp data, replace missing values with zeros, and apply the z-score and the min-max normalization. 
We use 70\% of the data for training, 10\% for validation, and the rest for testing.
Our pre-processing follows the approach used in DCRNN~\cite{li2018dcrnn}~\footnote{\url{https://github.com/liyaguang/DCRNN/tree/master/data}}.
We present a comprehensive experiment result with the METR-LA dataset, as the LA area's traffic conditions are more complicated than those in the Bay area~\cite{li2018dcrnn}. 
% In this paper, we present the comprehensive experimental results of just the METR-LA dataset, as the Bay area’s traffic conditions were more complicated than those in the LA area [11].

\subsection{Experimental Setup} 
\label{sec_settings}
\toolname predicts the speeds of the next 12 time steps from present time (five-minute intervals, one-hour in total) based on the input speeds of the previous $T=12$ time steps.
%12 time steps(5-minute interval, 1-hour in total) 
We train a four-layer spatio-temporal attention model ($H=4$, $d_{model}=128$). To reduce the discrepancy between the training and testing phase, We utilize an inverse sigmoid decay scheduled sampling~\cite{Bengio15scheduled}.

For optimization, we apply Adam-warmup optimizer~\cite{Vas2017transformer} and set the warmup step size and batch size as 4,000 and 20, respectively.
We use dropout (rate: $0.3$)~\cite{sariva2014dropout} on the inputs of every sub-layer and on the attention weights.
We initialize the parameters by using Xavier initialization~\cite{xavier10intial} and use a uniform distribution, $U(1,6)$, to initialize the weights of the diffusion prior. 
LINE~\cite{Tang2015LINE} is used for node embedding with dimension 64, which takes two minutes for training with 20 threads. 
We used a distance between two connected nodes and correlation using Vector Auto-Regression (VAR)~\cite{zivot2006vector} as an edge weight.

\begin{comment}
\subsection{Evaluation Metrics}

We evaluate the models using three metrics. For notation, we use $y_{i}^{*}$ as $i$-th example of predictions of the model, $y_{i}$ as a ground truth value of that example, and $|N_{y}|$ as the number of total samples. When $y_{i}$ is a missing value, we exclude $y_{i}$ from the evaluation set. 

The first metric is Mean Absolute Error (MAE), which is the average magnitude of difference between the prediction and the ground-truth:
\begin{equation}\label{eq.mae}
MAE =\frac{1}{|N_{y}|}\sum_{i=1}^{N_y}{|y_{i}^{*}-y_{i}|}.
\end{equation}

The second metric is Root Mean Squared Error(RMSE). While MAE has the same weight for all samples, RMSE increases the importance of samples with larger errors. It is defined as
\begin{equation}\label{eq.rmse}
RMSE =\frac{1}{|N_{y}|}\sum_{i=1}^{N_y}\sqrt{\left(y_{i}^{*}-y_{i}\right)^2}.  
\end{equation}

The last metric is Mean Absolute Percentage Error(MAPE). It is a variant of the MAE, but normalized by true observation. The characteristic of MAPE is that the value becomes huge when $y$ is a small value. It is defined as
\begin{equation}\label{eq.mape}
MAPE =\frac{100\%}{|N_{y}|}\sum_{i=1}^{N_y}{\frac{|y_{i}^{*}-y_{i}|}{y_{i}}}.
\end{equation}
\end{comment}

\subsection{Experimental Results}
\label{section.4.2}
we compare the performance of \toolname with six baseline models, including state-of-the-art deep learning models: [Graph Convolution based RNN (GCRNN), DCRNN~\cite{li2018dcrnn}, GaAN~\cite{Zhang2018gaan}, STGCN~\cite{yu2018spatio}, Graph WaveNet~\cite{Wu2019GraphWF} GMAN~\cite{zheng2019gman} and HyeperST~\cite{Pan2018HyperSTNetHF}]. 
%As we observed in our experiment that the performance of reproduced DCRNN on the METR-LA dataset is better than that in the original work~\cite{li2018dcrnn}, we use the reproduced result for comparison. 
%Other than DCRNN, we could not achieve better performance than the reported results, so we take the original results for comparison. 
We reproduce DCRNN~\footnote{\url{https://github.com/liyaguang/DCRNN/}}, Graph Wavenet~\footnote{\url{https://github.com/nnzhan/Graph-WaveNet}}, and GMAN~\footnote{\url{https://github.com/zhengchuanpan/GMAN}}, using the source codes and hyper parameters published by the authors.
When the source code is not available, we use the original results from the paper for comparison.
%\cb{To evaluate other baseline models, we report results by using official codes, which are published by authors, such as} DCRNN~\footnote{\url{https://github.com/liyaguang/DCRNN/}}, Graph Wavenet~\footnote{\url{https://github.com/nnzhan/Graph-WaveNet}} and GMAN~\footnote{\url{https://github.com/zhengchuanpan/GMAN}}. 
%\cb{If we are not able to access official codes, we take the original results from the paper for comparison. }
%When results are not available from the papers, we put `--'.
A few papers are not only unavailable to access official codes but also unpublished results in PEMS-Bay(Table~\ref{table.pems}).
%  a few paper is not only unavailable to access official codes but also unpublished results in PEMS-Bay.
In the experiment, we measure the accuracy of the models using absolute error (MAE), root mean squared error (RMSE), and mean absolute percentage error (MAPE).
The task is to predict the traffic speed 15, 30 and 60 minutes from present time. 
We also report average scores of three forecasting horizons on the dataset.

\begin{figure}[t]
\centering
\includegraphics[width=1.0\columnwidth]{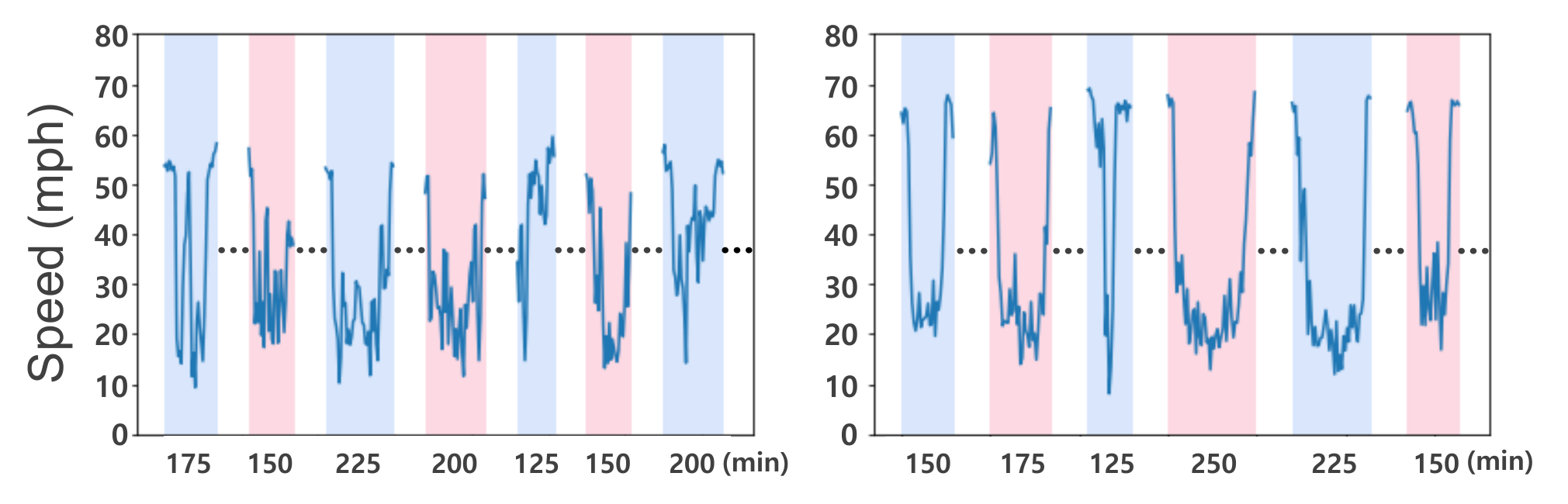}
%\vspace{-0.3cm}
\caption{Two examples of the intervals extracted by ruptures with the y-axis showing the speed and the x-axis showing the duration for each interval.}
\label{fig_congestion_interval}
%\vspace{-0.5cm}
\end{figure}

Table~\ref{table.metr} and Table~\ref{table.pems} show the experimental results of both datasets, where we observe that \toolname achieves state-of-the-art performance in the average scores. 
In particular, \toolname excels at predicting speed after 15 and 30 minutes. 
\toolname shows higher accuracy than GaAN, which is based on undirected spatial attention but neglects the proximity information. 
\toolname also achieves higher performances compared to DCRNN and GraphWavenet, demonstrating that our spatial and temporal attention mechanism is more effective than that of the two models for predicting short-term future sequences. 

We further evaluate \toolname from several perspectives.
First, we compare the forecasting performance of the models in separate time ranges. 
We do this because traffic congestion patterns in a city change dynamically over time. 
For example, some roads around residential areas are congested during regular rush hour periods, while those around industrial complexes are congested from late night to early morning~\cite{Lee19}.
Table~\ref{table:impededinterval} shows the experimental results for four time ranges (00:00--05:59, 06:00--11:59, 12:00--17:59, and 18:00--23:59) and we find that \toolname performs better than Graph WaveNet.
%We further evaluate \toolname in impeded interval such as rush hours and vehicle accidents. 
%To evaluate how \toolname adapts to speed changes, we first extract intervals of speeds from the data where road speeds change rapidly.

Secondly, to evaluate how \toolname adapts to speed changes during rush hour times, we extract intervals of speeds from the data where road speeds change rapidly.
We use ``ruptures"~\cite{Truong18}, an algorithm for computing changing points in a non-stationary signal.
Hyper-parameters of ruptures are 10, 1, and 6 for penalty value, jump, and minimum size, respectively.
Fig.~\ref{fig_congestion_interval} shows the extracted intervals of two example roads where the y-axis represents speed and the x-axis represents the sequence of intervals.
To find the intervals related to traffic congestion, we filter out intervals in which the slowest speeds are slower than 20 mph.
Table~\ref{table:impededinterval} shows the performance comparison between \toolname, Graph WaveNet and DCRNN, and we observe that our model captures the temporal dynamics of speed better than the others.

\begin{figure}[t]
\centering
\includegraphics[width=.8\columnwidth]{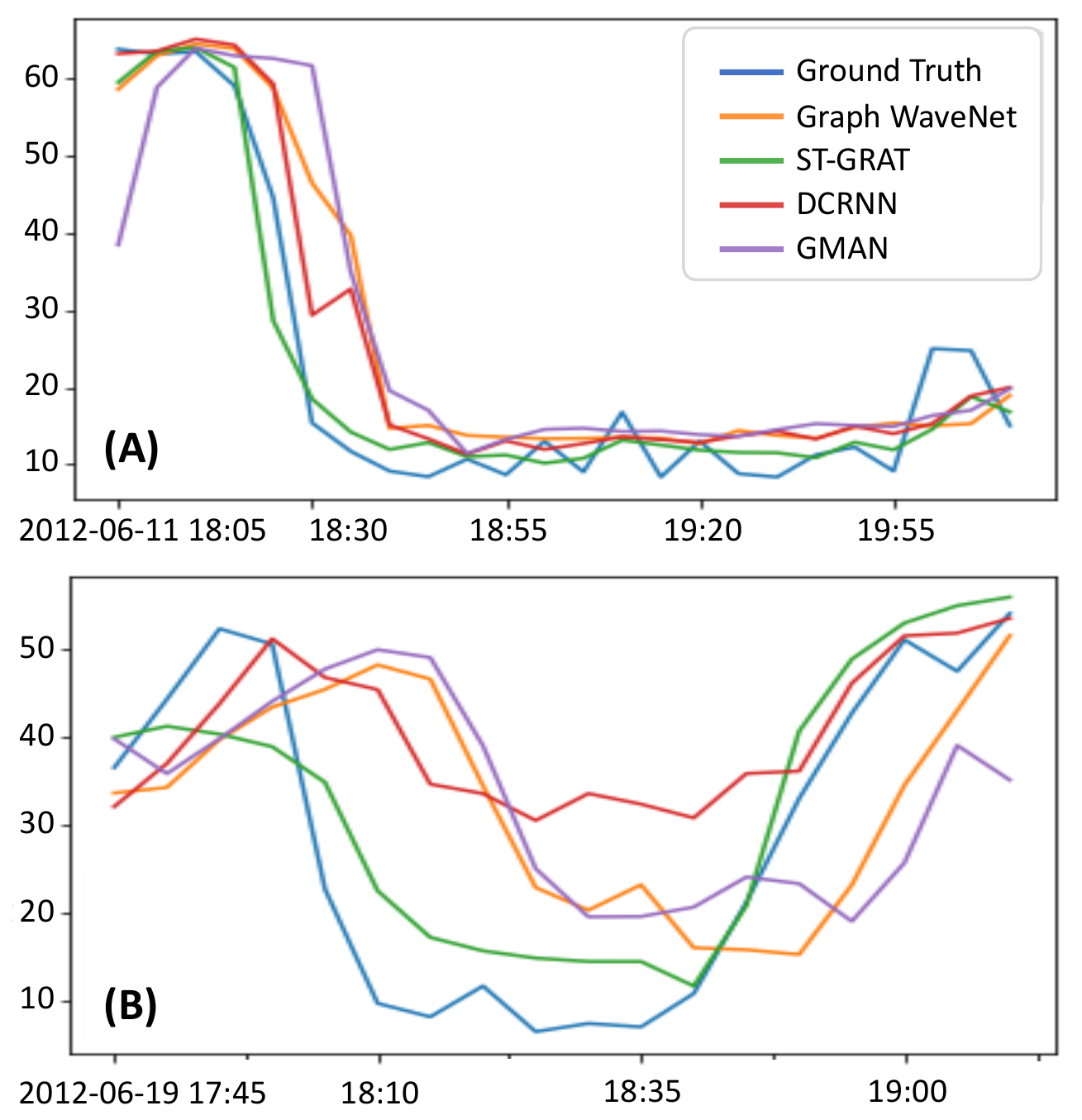}
%\vspace{-0.3cm}
\caption{Traffic speed prediction visualization in the impeded intervals.
ST-GRAT generates more accurate prediction and is robustness, especially when speeds abruptly change.}
\label{fig_linechart}
%\vspace{-0.5cm}
\end{figure}

Third, we visualize the traffic speed of each model on line charts to illustrates the traffic speed prediction patterns during impeded time intervals in the METR-LA dataset.
% Third, we visualized the traffic speed of each model on line charts to illustrate the traffic speed prediction patterns during the impeded time intervals in the METR-LA dataset.
%of 15 minute ahead on the METR-LA dataset in impeded interval by filtering ruptures. 
%better understand prediction of models in the impeded time intervals.
%The visualization illustrates traffic speed prediction of 15 minute ahead on the METR-LA dataset in impeded interval by filtering ruptures.
%The blue, orange, green, red, and purple represent ground truth, Graph WaveNet, ST-GRAT, DCRNN, and GMAN, respectively.
There is no time lag between ground truth and ST-GRAT prediction, while prediction traffic speed of other baseline models follow ground truth after reduced traffic speed as shown in Figure \ref{fig_linechart} (A).
The \toolname is more accurate with the abruptly changes and impeded interval than baseline models as shown in Figure \ref{fig_linechart} (B).
This is because ST-GRAT exploits the overall graph structure information and effectively encodes temporal dynamics.

\begin{table*}[]
\centering
\caption{Experimental results with different time ranges and intervals.}
\resizebox{\textwidth}{!}{
\begin{tabular}{l|lll|lll|lll|lll}
\hline
\multirow{2}{*}{Model / T} & \multicolumn{3}{l}{15 min} & \multicolumn{3}{l}{30 min} & \multicolumn{3}{l}{1 hour} & \multicolumn{3}{l}{Average} \\ \cline{2-13} 
                       & MAE    & RMSE    & MAPE    & MAE    & RMSE    & MAPE    & MAE    & RMSE    & MAPE    & MAE     & RMSE    & MAPE    \\ \hline
Our Model (00-06)                   & \textbf{2.37} & \textbf{3.57} & \textbf{4.30\%} & \textbf{2.41} & \textbf{3.67} & 4.44\% & 2.45 & 3.77 & \textbf{4.53\%} & \textbf{2.38} & \textbf{3.63} & \textbf{4.37\%} \\
Graph WaveNet (00-06)                & 2.41 & \textbf{3.57} & 4.38\% & 2.45 & 3.68 & 4.50\% & 2.52 & 3.75 & 4.61\% & 2.44 & 3.65 & 4.47\% \\
DCRNN (00-06) & 2.4 & 3.58 & 4.36\% & 2.44 & 3.72 & 4.52\% & 2.48 & 3.8 & 4.6\% & 2.43 & 3.68 & 4.47\% \\
GMAN (00-06) & 2.39 & 3.59 & 4.35\% & \textbf{2.41} & 3.69 & \textbf{4.43\%} & \textbf{2.43} & \textbf{3.75} & \textbf{4.48\%} & 2.40 & 3.66 & 4.40\% \\
\hline
Our Model (06-12)                   & \textbf{2.55} & \textbf{5.31} & \textbf{6.93\%} & \textbf{3.07} & \textbf{6.72} & \textbf{8.82\%} & \textbf{3.68} & 8.16 & \textbf{10.93\%} & \textbf{3.01} & \textbf{6.51} & \textbf{8.63\%} \\
Graph WaveNet (06-12)                & 2.67 & 5.43 & 7.55\% & 3.19 & 6.79 & 9.42\% & 3.73 & \textbf{8.01} & 11.28\% & 3.11 & 6.55 & 9.15\% \\
DCRNN (06-12) & 2.74 & 5.67 & 7.59\% & 3.28 & 7.16 & 9.6\% & 3.9 & 8.64 & 11.68\% & 3.22 & 6.95 & 9.35\%\\ 
GMAN (06-12) & 2.79 & 5.84 & 7.78\% & 3.22 & 7.00 & 9.27\% & 3.69 & 8.16 & 10.82\% & 3.17 & 6.83 & 9.09\%\\
\hline
Our Model (12-18)                   & \textbf{3.14} & \textbf{6.16} & \textbf{9.34\%} & \textbf{3.80} & 7.69 & \textbf{12.02\%} & 4.55 & 9.29 & 15.53\% & \textbf{3.72} & 7.47 & \textbf{11.88\%} \\
Graph WaveNet (12-18)                & 3.29 & 6.23 & 10.25\% & \textbf{3.80} & \textbf{7.49} & 12.41\% & \textbf{4.48} & \textbf{8.85} & 15.39\% & 3.77 & \textbf{7.31} & 12.28\% \\
DCRNN (12-18) & 3.33 & 6.43 & 10.11\% & 3.95 & 7.83 & 12.58\% & 4.66 & 9.33 & 15.66\% & 3.80 & 7.64 & 12.39\% \\
GMAN (12-18) & 3.48 & 6.87 & 10.78\% & 3.96 & 7.99 & 12.78\% & \textbf{4.48} & 9.12 & \textbf{15.26\%} & 3.90 & 7.81 & 12.65\%\\
\hline
Our Model (18-24)                   & \textbf{2.34} & \textbf{4.77} & \textbf{5.69\%} & \textbf{2.72} & \textbf{5.86} & \textbf{7.00\%} & 3.21 & 7.10 & \textbf{8.60\%} & \textbf{2.69} & 5.75 & \textbf{6.91\%} \\
Graph WaveNet (18-24)                & 2.41 & 4.83 & 6.09\% & 2.78 & 5.87 & 7.47\% & 3.19 & \textbf{6.90} & 8.99\% & 2.73 & \textbf{5.71} & 7.33\% \\
DCRNN (18-24) & 2.48 & 5.09 & 6.24\% & 2.85 & 6.19 & 7.79\% & 3.27 & 7.33 & 9.53\% & 2.81 & 6.04 & 7.66\% \\
GMAN (18-24) & 2.52 & 5.23 & 6.49\% & 2.82 & 6.16 & 7.78\% & \textbf{3.14} & 7.07 & 9.10\% & 2.78 & 6.03 & 7.65\% \\
\hline
\hline
Our Model (Impeded Interval)                   & \textbf{5.82} & \textbf{9.97} & \textbf{29.0\%} & \textbf{7.94} & \textbf{13.31} & \textbf{41.20\%} & \textbf{10.56} & 16.92 & \textbf{57.96\%} & \textbf{7.81} & \textbf{13.32} & \textbf{41.04\%} \\
Graph WaveNet (Impeded Interval)    & 6.45 & 10.69 & 33.19\% & 8.62 & 14.10 & 45.89\% & 11.14 & 17.65 & 62.79\% & 8.44 & 14.05 & 45.49\% \\

DCRNN (Impeded Interval)                & 6.44 & 10.41 & 33.75\% & 8.39 & 13.52 & 45.22\% & 10.71 & \textbf{16.76} & 60.93\% & 8.22 & 13.45 & 44.91\% \\
GMAN (Impeded Interval) & 6.94 & 11.55 & 36.05\% & 8.76 & 14.41 & 46.93\% & 10.74 & 17.22 & 60.38\% & 8.60 & 14.29 & 46.53\%
 \\
\hline
\end{tabular}
}
\label{table:impededinterval}
%\vspace{-0.5cm}
\end{table*}
\begin{table}[t]

\centering
\caption{Ablation study of \toolname with the METR-LA validation set. }
\resizebox{\columnwidth}{!}{\begin{tabular}{l|llllllll|l}
\hline
 & $L$ & $d_{model}$ & Embedding & $H$ & Range & Directed & Prior & Sentinel & MAE   \\ \hline
 \multirow{1}{*} Our model &4 & 128 & Distance & 4 & 2 & \ding{52} & \ding{52} & \ding{52} &2.80  \\ \hline

\multirow{2}{*}{(A)}  & 2 & & & & & & & & 3.02   \\ 
                     & 3 & & & & & & & & 2.98   \\\hline
\multirow{1}{*}{(B)}  &  & 64 &  & & & & & &  2.89   \\ \hline

\multirow{2}{*}{(C)}  &  &  & -- & & &  & & & 2.97    \\ 

                     &  & & Random & & & & & & 2.90   \\\hline
\multirow{2}{*}{(D)}  &  & & &2 & &  & & & 3.01   \\ 
                     &  & & &8& & & &  & 2.88 \\\hline
\multirow{1}{*}{(E)}  &  & & & & 1 & & & & 2.98 \\ \hline

\multirow{1}{*}{(F)}  &  & &  & &  &-- & &  & 2.87  \\ \hline
\multirow{1}{*}{(G)}  &  & &  & & &   & -- & &  2.85 \\ \hline 
\multirow{1}{*}{(H)} &  & &  & & &  & & -- &  2.89 \\ 
\hline

\end{tabular}}
\label{talbe.abl}
\end{table}

\subsection{Ablation Study}

\begin{comment}

\end{comment}

%\vspace{-0.01in}
\begin{comment}

\end{comment}

We conduct an ablation study to understand the impact of different hyper-parameters.
Table~\ref{talbe.abl} shows the experimental results where ${L}$,  $d_{model}$, and ${H}$ denote the number of layers, dimension of hidden state features, and number of heads, respectively.  
``Embedding'' and ``Range'' indicate different embedding methods and attention step sizes. 
If the range is one, it means that the model only attends nodes within a single step, such as directly connected nodes. 
If the range is two, it means the model attends nodes within two steps (i.e., directly linked nodes and their neighbors).
The hyphens denote excluded parameters. 
``Directed'' indicates that the input graph is a directed graph.
Finally, ``Prior'' and ``Sentinel'' indicate whether the model uses a diffusion prior and sentinel vector or not.
We show the average MAE prediction results from five-mins to one-hour measured by each evaluation metric.
Other settings are identical to those used in \autoref{sec_settings}.

%We conduct an ablation study to understand how different hyper-parameter settings influence our model. In Table~\ref{talbe.abl}, we fill in the table if we set a different hyper-parameter from the base model. The hyphen notation means we exclude the selected hyper-parameter.${L}$ indicates the number of layers, $d_{model}$ is the dimension of hidden state features, ``Embedding'' indicates the different embedding strategies, and ${H}$ is the number of heads.``Range'' is the indicator of attention step size. If the range is one, it means that model only attends nodes within a single step, such as directed connected nodes. If the range is two, it means the model attends nodes within two steps, which means directed connect nodes and its connected nodes. ``Directed'' shows whether the input graph is a directed or undirected graph. Finally, ``prior'' and ``sentinel'' indicate whether the model uses a diffusion prior and sentinel vector or not. We show the average MAE of prediction results from 5-mins to 1-hour measured by each evaluation metric. The rest of the settings are same as we describe in Experimental Settings. The results are reported in table ~\ref{talbe.abl}.

%The result of row (A) shows that the number of layers proportionally positively affects, because a small number of layers causes under-fitting. 

According to row (A), we observe that the number of layers proportionally affects the performance of our model. A deep model has better performance than a shallow model, because having a small number of layers causes the model to underfit. 
In row (B), we find that the dimension of the hidden vector also impacts the final performance. When the dimension of the hidden state vector increases, the key and query dimension also increases, allowing more information to be encoded. This motivates the model to better consider the elaborate relation between the query and key, leading to higher performance than a model with a small dimension of hidden vectors. Moreover, as we expected, we observe a positive correlation between the number of heads and the performance of the model in row (D). This is because as the number of heads increases, the model can capture more diverse aspects of spatial dependency than a model with a small number of heads. For example, if the model only has two heads, the first head might only consider the nodes that are experiencing traffic congestion while the second head only focuses on nodes that have light traffic.

Row (C) shows the results of different node embedding settings in comparison to our model, which utilize the distance between nodes as the node-embedding. We remove or add the embedding layer with different strategies as random initialization. Since the removed embedding layer and random initialization embedding does not have proximity information, both embedding layer models improperly perform attention, failing to attend more on close nodes than other distant nodes. By comparing different embedding settings, we observe that proximity information is hardly influencing the performance of our model.
% Row (C) shows the result of different node-embedding settings compared to our distance-based model.
% We removed or added the embedding layer with different strategies as random initialization.

%In the small sampled attention ($Range=1$), in order to predict the distant future, we should refer to nodes that are now far away but will affect the future.

From row (E), we show that the performance is also affected by the range of neighbor nodes that carry out the spatial attention process. The performance of $Range=2$ is better than that of $Range=1$ because the model with $Range=2$ considers more neighbor nodes than the model with $Range=1$. However, as the number of $Range$ increases, the computation cost of nodes (i.e., scaled-dot product operation) also increases.

Row (F) shows the effect of directed attention (inflow and outflow attention). In undirected attention, the spatial attention needs to compare the similarity of the query and key to all in-coming or out-going nodes from a particular node. As a result, even though undirected attention takes more nodes into consideration, its performance is lower than our model which uses directed attention. This experiment shows that our directed attention, which splits the heads into inflow and outflow, is suitable to solve traffic problems. Row (G) indicates whether the model applies the diffusion prior. The base model performance is better than the model that does not apply the diffusion prior. This is because the model adjusts the attention logits by utilizing a diffusion prior.
 
Row (H) shows the effect of a sentinel key-value vector, which is described in the spatial attention section. We find that the model that uses the sentinel vectors is better than the model that does not use sentinel vectors. This result indicates that the sentinel vectors help the model to improve robustness in the traffic prediction by avoiding unnecessary attentions.

\section{Qualitative Analysis}
\begin{figure*}[t]
\centering
\includegraphics[width=\linewidth]{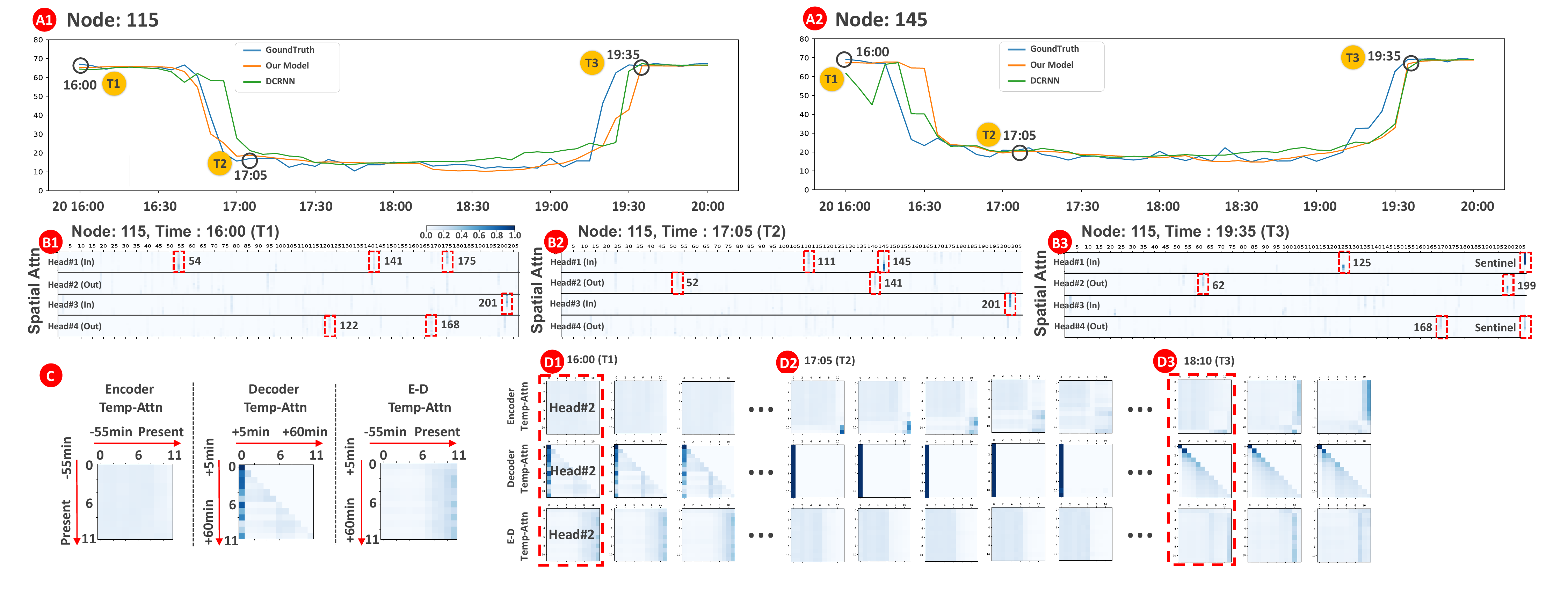}
\caption{(A) Two line charts of the current node and neighbor node, (B) Spatial-attention heatmaps in different time steps, and (C) Temporal-attention heatmaps in different time steps at METR-LA dataset in impeded interval (2012/06/20 16:05–2012/06/20 20:10).
}
\label{fig_qualitative_analysis}
\end{figure*}
%In this section, we describe our model's prediction process to show how deep learning models make decisions by analyzing the attention dynamics.
In this section, we describe how \toolname captures spatio-temporal dependencies within a road network.
Fig.~\ref{fig_qualitative_analysis} (A1) shows speed changes of Node 115, where the x- and y-axes represent time and speed (mph), respectively.
As shown in Fig.~\ref{fig_qualitative_analysis} (A1), this road was congested from 16:50 and congestion was alleviated at approximately 19:35 (i.e., a typical congestion pattern in the evenings).
The heatmaps of Fig.~\ref{fig_qualitative_analysis} (B1--B3) provide information on spatial attention of the last layer at different times (B1--T1, B2--T2, B3--T3). 
The y-axis at each head (in-out-in-out degrees) represents 12 time sequences, and the x-axis indicates the 207 nodes and the sentinel vector (last column). 

First, we find that \toolname gave more attention to six nodes (54, 122, 141, 168, 175, and 201) than others in light traffic (T1 in Fig.~\ref{fig_qualitative_analysis} (A1)). 
%Note that Node 115 always received attention. 
Then, as time passes from T1 to T2, Nodes 54 and 175 gradually received less attention, and Nodes 52, 111, 145, and 201 gained more attention, as shown in Fig.~\ref{fig_qualitative_analysis}(B2). 
%One question we have at this point is that why Node 145 was still in the attention list as time proceeded, while other nodes gradually disappeared from the attention list.
A notable observation is that \toolname attended to Node 145 and Node 201 (i.e., dark blue bars) more than other nodes.
%In addition, how important is 145 and why Node 111 chose to keep Node 145 in its attention list, regardless of congestion conditions.
% To investigate, we first review Node 145’s speed records (Fig. 4, A2) and find Node 145’s speed tended to precede that of Node 115 by about 30 minutes.
We review speed records of Node 145 (Fig.~\ref{fig_qualitative_analysis}, A2) due to its strong attention and find that the speed of Node 145 tended to precede that of Node 115 by about 30 minutes (Fig.~\ref{fig_qualitative_analysis} (A2)).
%In addition, when we compute correlations between Node 115 and other nodes, Node 145 is strongly correlated to Node 115 (correlation: 0.64), while other nodes show lower correlations (e.g., 0.344 with Node 62).
%Checking correlations between two nodes based on Vector Auto-Regression, we find that speed pattern of Node 115 was highly correlated with that of Node 145 and 201 (correlation: 0.59 and 0.56, respectively), while it was less correlated with other nodes (e.g., correlation: 0.48 with Node 52). 
In checking the correlations between pairs of nodes based on the VAR, we find that the speed pattern of Node 115 was highly correlated with that of Nodes 145 and 201 (0.59 and 0.56, respectively), while it was less correlated with other nodes (e.g., 0.48 with Node 52).
An interesting point here is that the distances from the two nodes to Node 115 are different (0.9 miles from Node 145 and 7.64 miles from Node 201).
This result shows that \toolname learned dynamically changing spatial dependencies along with the graph structure information such as correlations and distances.
%An interesting point here is that the distances from the two nodes to Node 115 are different--0.9 miles from Node 115 to Node 145 and 7.64 miles from Node 115 to Node 201.
%We believe this is the result of learning spatial relationships--\toolname chose to attend to two highly correlated nodes at different distances, when the road condition changed from light to heavy traffic.
% 
%One interesting point in the correlation analysis is that Node 54 also had a high correlation with Node 118 (correlation: 0.59), but this node did not have much attention from Node 118 in impeded conditions.

\begin{figure*}[t]
\centering
\includegraphics[width=1.0\linewidth]{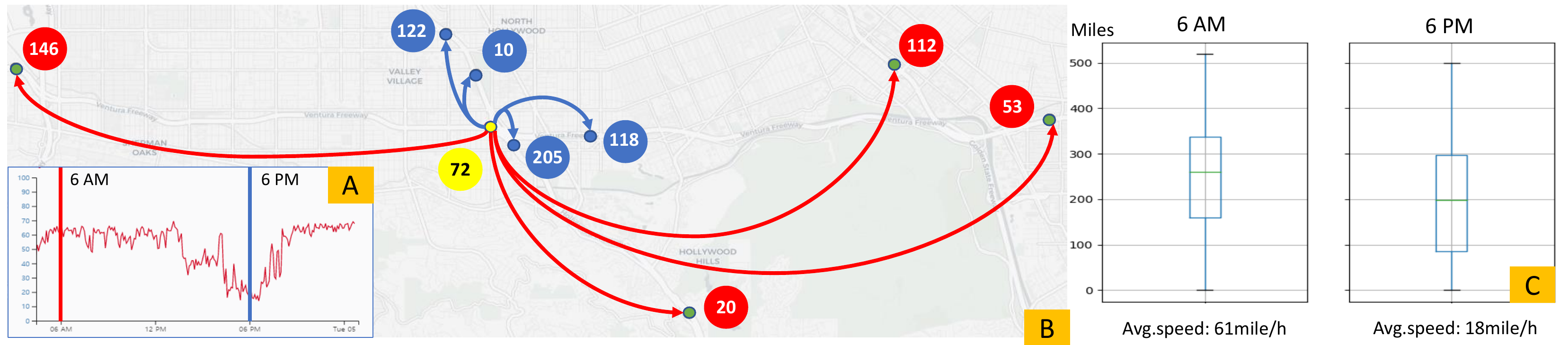}
%\vspace{-0.3cm}
\caption{An example spatio-temporal dependency modeling based on node speeds and distance between query and key nodes. (A) Node 72's speed trend; (B) nodes attended by Node 72 (yellow). Red and blue nodes are attended at 6am (unimpeded) and 6pm (impeded), respectively; and (C) an overall distribution of distances between the query nodes and corresponding key nodes. Node 72 attended the nodes closed to itself as the road became congested (p-value: $\ll 0.01$).}
\label{fig_spataio_temporal}
%\vspace{-0.5cm}
\end{figure*}
% An example of spatio-temporal dependency modeling based on node speeds and distance between query and key nodes: (A) Node 72’s speed trend; (B) nodes attended by Node 72 (yellow), red and blue nodes are attended at 6 am (unimpeded) and 6 pm (impeded), respectively; and (C) an overall distribution of distances between the query nodes and corresponding key nodes. Node 72 attended to the nodes close to itself as the road became congested (p-value < 0.01).
When the traffic congestion was alleviated at 19:35 (Fig.~\ref{fig_qualitative_analysis}, T3), \toolname attended to its sentinel vector of Head\#1 (In) and Head\#4 (Out), as shown in Fig.~\ref{fig_qualitative_analysis} (B3).
%This implies that \toolname decided to focus on the spatial dependency captured from 18:45 to 19:05 (i.e., index 0 to index 7 of the sentinel vector), where the speed started increasing.
This implies that \toolname decided to utilize the existing features extracted from 18:45 to 19:05 by attending sentinel vector. 
%Index 0 (top) means the time step of 55 minutes ago and index 11 (bottom) represents the present time step. 
\toolname also attended to Node 125 (correlation: 0.43) for updating spatial dependency from a neighboring road (2.7 miles away from Node 115). 
We see a similar behavior in Head\#4 (Out)--\toolname focused on the sentinel vector, while attending to a distant Node 168 (correlation: 0.59, 7.8 miles away from Node 115). 
In the spatial attention, \toolname dynamically uses new features from spatially correlated nodes or preserves existing features.

Next, we analyze the second head of the first layer in the encoder, decoder, and encoder-decoder (E-D) temporal attention (Fig.~\ref{fig_qualitative_analysis} (D1)). 
Note that Fig.~\ref{fig_qualitative_analysis}(C) shows how the input and output sequences are mapped with each axis of the attention heatmaps in D1--D3. 
We see from Fig.~\ref{fig_qualitative_analysis} D1 that when Node 115 was not congested (T1), the  temporal attention of the encoder and the decoder were equally spread out across all time steps.
%decoders' temporal attention focused on the first column, which implies that \toolname mainly attended to Node 115's current traffic condition. 
However, it is interesting that the temporal attention of the encoder gradually became divided into two regions (top-left and bottom-right) as the road became congested, as shown in a series of the heatmaps in Fig.~\ref{fig_qualitative_analysis} (D2).  
%Here, the top-left region represents past unimpeded conditions' information, while the other region means current impeded conditions' information.
Here, the top-left region represents the information of past unimpeded conditions, while the other region contains information on the current impeded conditions.
%\toolname needed to consider two possible future directions at the same time--one with static congestion and another with a changing condition. 
We believe this divided attention is reasonable, as \toolname needed to consider two possible future directions at the same time--one with static congestion and another with a changing condition, possibly with a recovering speed. 
It is notable that the first column of the temporal attention of the decoder became darker (D2), which implies that \toolname attended to recent information in the impeded condition. 
Overall, \toolname uses the attention module in an adaptive manner with evolving traffic conditions for effectively capturing the temporal dynamics of traffic. 

\subsection{Attention Patterns in Different Traffic Conditions}
In order to accurately model the spatial dependencies among roads, models need to dynamically adjust spatial correlations based on different traffic conditions (e.g., impeded condition) and the graph structure information, such as the connectivity, directions, and distances between nodes.
However, previous models do not fully utilize the graph structure information in the spatial modeling~\cite{Zhang2018gaan,zheng2019gman}.
%For example, GaAN does not cares the connectivity and directions of nodes~\cite{zheng2019gman}, and GaAN only takes the connectivity among nodes~\cite{Zhang2018gaan}.
In cotrast, \toolname incorporates graph structure features in the spatial attention by using diffusion priors and distance embedding and directed heads. 
In this section, we describe how \toolname dynamically adjusts spatial correlations based on road conditions. 
%On the other hands, ST-GRAT adjusts spatial attention weights based on the speed pattern and the graph structure information between the query node and the key nodes by using diffusion priors, distance embedding and directed heads 

Fig.~\ref{fig_spataio_temporal} describes an example attention pattern with  Node 72. 
As shown in Fig.~\ref{fig_spataio_temporal} (A), Node 72 is not congested early in 6 am (average speed: 61 miles/h), while it is impeded at 6 pm (average speed: 18 miles/h). 
Fig.~\ref{fig_spataio_temporal} (B) shows that during the impeded traffic condition (6 pm), Node 72's key nodes (blue dots) are much closer to the query Node 72 when compare to the unimpeded traffic condition (6 am) key nodes (red dots).
Note that this is an magnified map, and key nodes with attention weights less than 0.1 are filtered out. 
Fig.~\ref{fig_spataio_temporal} (C) shows the overall distance distribution of key nodes of 144 query nodes on June 4, 2012, 6 am and 6 pm. 
We choose the 144 query nodes that have less than 30 miles/h at the time and pick the longest distance between query and key nodes. 
We find that query nodes use the information of key nodes at significantly closer distances (average distance: 326.4 miles, STD: 206.7) in the impeded condition, compared to the unimpeded condition(average distance: 414.3, std: 186,3), according to the independent two sample t-test result (t[143]=3.3, p-value$<$0.01). 
%This patterns show that model dynamically controls the attention weights depend on the speed and distance of query nodes between key nodes.
% 이 결과는 
This result shows that \toolname predicts road speeds by considering the road speeds, distances between query and key nodes, and the road network, which leads to a better prediction performance than the existing models. 

%(average distance: 326.4 miles, STD: 206.7) (average distance: 414.3, std: 186,3).

%6AM (M = 414.3, SD = 186.3), 6PM (M = 326.4, SD = 206.7),
%t(143) = 3.3, p = .001. 

%It demonstrates that the model tends to select nodes depending on the query node speeds and distance with key nodes. 
%To identify this pattern generally shows 
%#For instance, as shown in %Fig.~\ref{fig_spataio_temporal} at C, overall max distances of attended key nodes from the query nodes become much far away when the speed is faster where attended key nodes defines as the attention weight of its nodes is higher than 0.1.
%For the detail, Fig~\ref{fig_spataio_temporal}(B)  shows that how exactly the model adjusts the reference nodes based on the speed and the distance.

%Furthermore, the bottom histogram in the figure shows that when the speed is faster, the overall number of attended key nodes becomes bigger, since the reference networks which the model considers also becomes bigger based on the distance and speed.

% 기존의 attention 기반의 연구들은 거리를 고려하지 않고 attention을 하지 않았다. gman의 경우 연결된 노드와 방향 정보를, ganan은 노드간의 연결성만을 고려한 것을 확인 할 수 있다. 하지만 우리의 모델에서는 주변상황과 자기자신 상태에 맞게 참조하는 도로가 달라질때 도로망정보 (연결성, 거리 방향등을 고려한다.) 
% x 에제와 같이 rush hour 시간이 아닐때에는 속도가 원활하므로 주변의 먼 노드까지 참조하는 것을 확인 할 수 있는데. 막히기 시작하는 rush hour시간에는 도로망 정보를 참조하여, 주변의 범위의 도로들을 주로 참조하는 것을 확인 할 수 있다. 이는 모델에 도로망 정보를 고려하도록 하여 모델이 필요하다면 거리 정보를 활용하여 attnetion 값을 조정함으로써 모델의 성능을 향상 시킨 것으로 보인다.  

\subsection{Analysis of Sentinel Vectors} 
\begin{figure}[t]
\centering
\includegraphics[width=\linewidth]{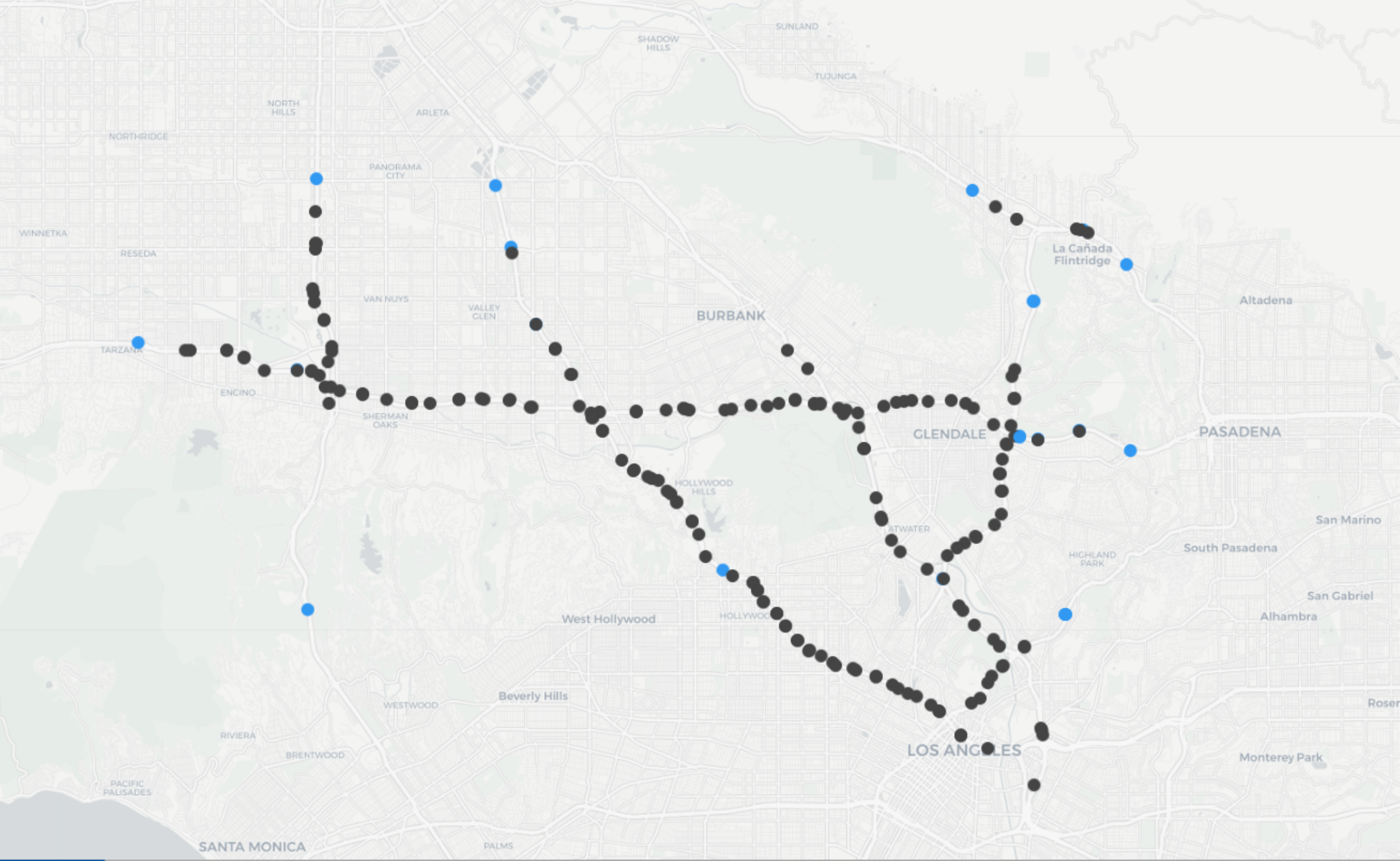}
\caption{Highly affected nodes by sentinel vectors in METR-LA data. We visualize a sensor distribution of METR-LA dataset. Especially, we mark highly affected nodes by sentinel vector as blue color where sentinel attention weight($\alpha_{i,s}$) of the blue node has more than $0.35$.}
\label{fig_qualitative_analysis_3}
\end{figure}

To analyze how sentinel vectors work in the speed prediction tasks, we investigate the nodes that extensively utilize the vectors. 
For this, we first choose the nodes that have a sentinel attention weight ($\alpha_{i,s}$) higher than 0.35 (the upper 10\% of the entire nodes), as shown in Fig.~\ref{fig_qualitative_analysis_3} as blue nodes.  
We then find it interesting that the nodes with high sentinel weights tend to be locate on the outskirts of the city. 
We also observe that they do not have many neighbor nodes compared to those in other areas (e.g., at the center of the city). 
Thus they do not have a sufficient pool of nodes with high spatial correlations for speed prediction. 
This means if a model is forced to encode new information based on spatial encoding, it is highly possible that it uses information of few other nodes, which may not be helpful for prediction. 
We also think that this attention strategy could negatively affect the performance of neighbor nodes that use this node as a key node. 
%The most of nodes are located on the out-side of the graph. 
%In METR-LA dataset, nodes of the out-side of the graph absent some of the highly correlated nodes.
%In other words, these nodes when predicting future speeds.
%These nodes do not have a few spatial correlation with given nodes, when predicting future speeds. 
%In this case, when we force to retrieve new information in spatial modeling, the model extracts unnecessary information. 
%Moreover, this pattern negatively affects its neighbor nodes, which attend nodes of the out-side. 

%Instead of attending unrelated nodes,
To avoid such situations, \toolname utilizes the sentinel vectors that contain features acquired in previous steps. 
%Owing to these characteristics, our model encodes refined features from given inputs by removing unnecessary attention.
% we compared ST-GRAT to its version without the vectors (ST-GRAT-NoSentinelVector) and found
To measure the effects of the vectors in such nodes, we compare \toolname  to its version without the sentinel vectors (\toolname-NoSentinelVector) and find that \toolname shows higher performance on the  METR-LA test data (\toolname: 2.27, \toolname-NoSentinelVector: 2.34).
As such, we believe that \toolname improves the prediction performance by leveraging the sentinel vectors. 
%our model with a model, which is removed sentinel vectors. 
%In these nodes, the without sentinel vector model and our model shows the average MAEs along the 12 sequences are 2.27 and 2.34 on METR-LA test data, respectively. 
%It demonstrate the effectiveness of the sentinel vectors in nodes of out-side. 
%Unlike previous models, \toolname utilizes the sentinel vectors on the out-side of road network, so the sentinel vector leverages the performances in the abrupt change point.

% sentinel vector를 분석하기 위해, 우리는 모델이 sentinel vecotr에 가장 많이 영향 받은 노드들을 확인해보고자 했다. 
% 이를 분석하기 위해 metr-la 데이터 셋에서 모델에서 평균 0.3 이상의 sentinel vector를 보는 도로들을 확인해보았다. 
% x번 figure 를 보면, 거의 대부분의 입력에 대해 센티넬 백터를 가장 많이 본 도로들을 추출해보면 주어진 도로들에서 외곽도로 인것을 확인해 볼 수 있었다. metr-la 데이터 셋에서 실험을 위해 전체도로중에서 일부 노드를 sampling 했기 때문에 외곽도로는 주변에 연결된 도로들이 일부가 없는 노드들이다.즉, 이러한 노드의 경우 미래 속도를 예측할때 공간적으로 주변의 참조할 정보가 없는 경우가 있다. 이러한 경우에서 attnetion 을 sum을 1로 하도록 강제하게 되면, 불필요한 정보를 참조하게 된다.  하지만 우리 모델에서는 불필요한 어텐션을 하기 보다는, 자기자신의 historical 한 pattern정보를 유지하도록 센티넬 벡터를 봄으로써 필요한 정도만 어텐션 하게 된다. 그렇기 때문에 불필요한 다른 노드들을 보기보다는 이전레이어의 historical pattern정보를 유지하는 것을 확인 해 볼 수 있다. 

% 추가적으로 sentinel vector를 많이 본 예측 sequence를 보면 주로 rush hour나 historical하게 주기적인 영향을 받는 경우라는 것을 확인 할 수 있다. 위의 결과를 통해 이전레이어의 temproal feature를 가져오거나 주변에 spatial 정보를 보지 않고 예측을 하는게 좋은 경우 모델은 sentinel vector를 통해서 불필요 한 attention을 하지 않는 것을 확인 할 수 있다.

\begin{table}[t]
\caption{The computation times on the METR-LA dataset.}
\centering
\resizebox{\columnwidth}{!}{\begin{tabular}{l|lllll}
\hline
                                          Computation Time& DCRNN & GaAN & Graph Wavenet   & GMAN & \toolname  \\ 
\hline
 Training (s/epoch) & 504.4& 1461.4 &203.89 & 552.1&341.7\\
                Inference (s) & 34.0 & 131.10& 8.42 & 50.02 &48.67 \\

\hline

\end{tabular}}
\label{table.compu}
%\vspace{-0.5cm}
\end{table}

\subsection{Computation Time}
In this section, we report the computation costs  of the models with the METR-LA dataset, as shown in Table~\ref{table.compu}.
In terms of the training time, we find that \toolname is faster ($O(1)$) than RNN-based models, such as DCRNN and GaAN which have $O(N)$ time complexity~\cite{Vas2017transformer}.
Comparing  \toolname to the attention models, we find \toolname  is about four and three times faster than GaAN in the training and inference stages, respectively.
\toolname is also faster that GMAN which needs to consider more roads, as it does not uses the connectivity information between nodes.
The Graph WaveNet performs best because it is a non-autoregressive model.
Overall, \toolname is the second best model in terms of the training time cost, and shows an average performance in terms of the inference time. 
However, \toolname is more robust than other models in the impeded conditions  (Table~\ref{table:impededinterval}), which are more difficult to predict. 
Also, \toolname provides interpretability (Fig.~\ref{fig_qualitative_analysis}), which is an additional advantage over previous black-box models, such as GCNN-based models. 
%\vspace{-0.12in}

%We report the computation times for comparing spatio-temporal models on the METR-LA dataset in Table.~\ref{table.compu}. 
%\cb{%In order to report the computation time, we utilize official codes which are aforementioned at Section.4.2.
%In the case of training time, our method is faster than RNN-based models, such as DCRNN and GaAN. 
%The complexity of sequential operations of \toolname is only $O(1)$, while that of the RNN-based models is $O(N)$~\cite{Vas2017transformer}. In attention model cases, \toolname is almost four times and three times faster than the GaAN in the training and inference stages, respectively. GMAN slightly slower than our model due to the reason that this model does not reflect connectivity of nodes.  The graph wavenet performs best because it is a non-autoregressive model.}

\section{Conclusions}
In this work, we presented \toolname with a novel spatial and temporal attention for accurate traffic speed prediction.
Spatial attention captures the spatial correlation among roads, utilizing graph structure information, while temporal attention captures the temporal dynamics of the road network by directly attending to features in long sequences.  % provided by node embedding, diffusion prior, the additional spatial sentinel key-value pair and directed attention heads, 
\toolname achieves the state-of-the-art performance compared to existing methods on the METR-LA and the PEMS-BAY datasets, especially when speeds dynamically change in the short-term forecasting.
Lastly, we visualize when and where \toolname attends when making predictions during traffic congestion.
As future work, we plan to conduct further experiments using \toolname with different spatio-temporal domains and datasets, such as air quality data.

\section*{Acknowledgment}
This work was supported by Institute of Information \& communications Technology Planning \& Evaluation(IITP) grant funded by the Korea government(MSIT)--No.20200013360011001, Artificial Intelligence graduate school support(UNIST), and No.2018-0-00219, Space-time complex artificial intelligence blue-green algae prediction technology based on direct-readable water quality complex sensor and hyperspectral image. Sungahn Ko (UNIST) and Jaegul Choo (KAIST) are the corresponding authors.

\bibliographystyle{ACM-Reference-Format}
\bibliography{sample-base}

%%% -*-BibTeX-*-
%%% Do NOT edit. File created by BibTeX with style
%%% ACM-Reference-Format-Journals [18-Jan-2012].

\begin{thebibliography}{31}

%%% ====================================================================
%%% NOTE TO THE USER: you can override these defaults by providing
%%% customized versions of any of these macros before the \bibliography
%%% command.  Each of them MUST provide its own final punctuation,
%%% except for \shownote{}, \showDOI{}, and \showURL{}.  The latter two
%%% do not use final punctuation, in order to avoid confusing it with
%%% the Web address.
%%%
%%% To suppress output of a particular field, define its macro to expand
%%% to an empty string, or better, \unskip, like this:
%%%
%%% \newcommand{\showDOI}[1]{\unskip}   % LaTeX syntax
%%%
%%% \def \showDOI #1{\unskip}           % plain TeX syntax
%%%
%%% ====================================================================

\ifx \showCODEN    \undefined \def \showCODEN     #1{\unskip}     \fi
\ifx \showDOI      \undefined \def \showDOI       #1{#1}\fi
\ifx \showISBNx    \undefined \def \showISBNx     #1{\unskip}     \fi
\ifx \showISBNxiii \undefined \def \showISBNxiii  #1{\unskip}     \fi
\ifx \showISSN     \undefined \def \showISSN      #1{\unskip}     \fi
\ifx \showLCCN     \undefined \def \showLCCN      #1{\unskip}     \fi
\ifx \shownote     \undefined \def \shownote      #1{#1}          \fi
\ifx \showarticletitle \undefined \def \showarticletitle #1{#1}   \fi
\ifx \showURL      \undefined \def \showURL       {\relax}        \fi
% The following commands are used for tagged output and should be
% invisible to TeX
\providecommand\bibfield[2]{#2}
\providecommand\bibinfo[2]{#2}
\providecommand\natexlab[1]{#1}
\providecommand\showeprint[2][]{arXiv:#2}

\bibitem[\protect\citeauthoryear{Atwood and Towsley}{Atwood and
  Towsley}{2016}]%
        {atwood2016difconv}
\bibfield{author}{\bibinfo{person}{James Atwood} {and}
  \bibinfo{person}{Donald~F. Towsley}.} \bibinfo{year}{2016}\natexlab{}.
\newblock \showarticletitle{Diffusion-Convolutional Neural Networks}. In
  \bibinfo{booktitle}{\emph{Proc. the Advances in Neural Information Processing
  Systems (NIPS)}}.
\newblock


\bibitem[\protect\citeauthoryear{Ba, Kiros, and Hinton}{Ba
  et~al\mbox{.}}{2016}]%
        {Ba2016LayerN}
\bibfield{author}{\bibinfo{person}{Jimmy Ba}, \bibinfo{person}{Ryan Kiros},
  {and} \bibinfo{person}{Geoffrey~E. Hinton}.} \bibinfo{year}{2016}\natexlab{}.
\newblock \showarticletitle{Layer Normalization}.
\newblock   \bibinfo{volume}{abs/1607.06450} (\bibinfo{year}{2016}).
\newblock


\bibitem[\protect\citeauthoryear{Bengio, Vinyals, Jaitly, and Shazeer}{Bengio
  et~al\mbox{.}}{2015}]%
        {Bengio15scheduled}
\bibfield{author}{\bibinfo{person}{Samy Bengio}, \bibinfo{person}{Oriol
  Vinyals}, \bibinfo{person}{Navdeep Jaitly}, {and} \bibinfo{person}{Noam
  Shazeer}.} \bibinfo{year}{2015}\natexlab{}.
\newblock \showarticletitle{Scheduled sampling for sequence prediction with
  recurrent neural networks}. In \bibinfo{booktitle}{\emph{Proc. the Advances
  in Neural Information Processing Systems (NIPS)}}.
\newblock


\bibitem[\protect\citeauthoryear{Cho, Bahdanau, Farhadi, and Bengio}{Cho
  et~al\mbox{.}}{2014}]%
        {bahdnamu2014seq2attn}
\bibfield{author}{\bibinfo{person}{Kyunghyun Cho}, \bibinfo{person}{Dzmitry
  Bahdanau}, \bibinfo{person}{Farhadi}, {and} \bibinfo{person}{Yoshua Bengio}.}
  \bibinfo{year}{2014}\natexlab{}.
\newblock \showarticletitle{Neural Machine Translation by Jointly Learning to
  Align and Translate}.
\newblock \bibinfo{journal}{\emph{Proc. the International Conference on
  Learning Representations (ICLR)}} (\bibinfo{year}{2014}).
\newblock


\bibitem[\protect\citeauthoryear{Devlin, Chang, Lee, and Toutanova}{Devlin
  et~al\mbox{.}}{2019}]%
        {jacob2018bert}
\bibfield{author}{\bibinfo{person}{Jacob Devlin}, \bibinfo{person}{Ming-Wei
  Chang}, \bibinfo{person}{Kenton Lee}, {and} \bibinfo{person}{Kristina
  Toutanova}.} \bibinfo{year}{2019}\natexlab{}.
\newblock \showarticletitle{BERT: Pre-training of Deep Bidirectional
  Transformers for Language Understanding}. In \bibinfo{booktitle}{\emph{Proc.
  the Annual Meeting of the Association for Computational Linguistics (ACL)}}.
\newblock


\bibitem[\protect\citeauthoryear{Glorot and Bengio}{Glorot and Bengio}{2010}]%
        {xavier10intial}
\bibfield{author}{\bibinfo{person}{Xavier Glorot} {and} \bibinfo{person}{Yoshua
  Bengio}.} \bibinfo{year}{2010}\natexlab{}.
\newblock \showarticletitle{Understanding the difficulty of training deep
  feedforward neural networks}. In \bibinfo{booktitle}{\emph{Proceedings of the
  Thirteenth International Conference on Artificial Intelligence and
  Statistics}} \emph{(\bibinfo{series}{Proceedings of Machine Learning
  Research})}, Vol.~\bibinfo{volume}{9}. \bibinfo{publisher}{PMLR},
  \bibinfo{pages}{249--256}.
\newblock


\bibitem[\protect\citeauthoryear{Hendrycks and Gimpel}{Hendrycks and
  Gimpel}{2016}]%
        {hendrycks2016gelu}
\bibfield{author}{\bibinfo{person}{Dan Hendrycks} {and} \bibinfo{person}{Kevin
  Gimpel}.} \bibinfo{year}{2016}\natexlab{}.
\newblock \showarticletitle{Gaussian Error Linear Units (GELUs)}.
\newblock \bibinfo{journal}{\emph{arXiv:1606.08415}} (\bibinfo{year}{2016}).
\newblock


\bibitem[\protect\citeauthoryear{Hochreiter and Schmidhuber}{Hochreiter and
  Schmidhuber}{1997}]%
        {hochreiter1997lstm}
\bibfield{author}{\bibinfo{person}{Sepp Hochreiter} {and}
  \bibinfo{person}{J{\"u}rgen Schmidhuber}.} \bibinfo{year}{1997}\natexlab{}.
\newblock \showarticletitle{Long short-term memory}.
\newblock \bibinfo{journal}{\emph{Neural computation}} \bibinfo{volume}{9},
  \bibinfo{number}{8} (\bibinfo{year}{1997}), \bibinfo{pages}{1735--1780}.
\newblock


\bibitem[\protect\citeauthoryear{Kipf and Welling}{Kipf and Welling}{2016}]%
        {kipf2016semi}
\bibfield{author}{\bibinfo{person}{Thomas~N Kipf} {and} \bibinfo{person}{Max
  Welling}.} \bibinfo{year}{2016}\natexlab{}.
\newblock \showarticletitle{Semi-Supervised Classification with Graph
  Convolutional Networks}.
\newblock \bibinfo{journal}{\emph{Proc. the International Conference on
  Learning Representations (ICLR)}} (\bibinfo{year}{2016}).
\newblock


\bibitem[\protect\citeauthoryear{Lee, Kim, Jin, Kim, Maciejewski, Ebert, and
  Ko}{Lee et~al\mbox{.}}{2019}]%
        {Lee19}
\bibfield{author}{\bibinfo{person}{Chunggi Lee}, \bibinfo{person}{Yeonjun Kim},
  \bibinfo{person}{Seungmin Jin}, \bibinfo{person}{Dongmin Kim},
  \bibinfo{person}{Ross Maciejewski}, \bibinfo{person}{David Ebert}, {and}
  \bibinfo{person}{Sungahn Ko}.} \bibinfo{year}{2019}\natexlab{}.
\newblock \showarticletitle{A Visual Analytics System for Exploring,
  Monitoring, and Forecasting Road Traffic Congestion}.
\newblock \bibinfo{journal}{\emph{IEEE Transactions on Visualization and
  Computer}} (\bibinfo{year}{2019}).
\newblock


\bibitem[\protect\citeauthoryear{Li, Yu, Shahabi, and Liu}{Li
  et~al\mbox{.}}{2018}]%
        {li2018dcrnn}
\bibfield{author}{\bibinfo{person}{Yaguang Li}, \bibinfo{person}{Rose Yu},
  \bibinfo{person}{Cyrus Shahabi}, {and} \bibinfo{person}{Yan Liu}.}
  \bibinfo{year}{2018}\natexlab{}.
\newblock \showarticletitle{Diffusion Convolutional Recurrent Neural Network:
  Data-Driven Traffic Forecasting}.
\newblock \bibinfo{journal}{\emph{Proc. International Conference on Learning
  Representations (ICLR)}}.
\newblock


\bibitem[\protect\citeauthoryear{Lin, Feng, Santos, Yu, Xiang, Zhou, and
  Bengio}{Lin et~al\mbox{.}}{2017}]%
        {lin2017structured}
\bibfield{author}{\bibinfo{person}{Zhouhan Lin}, \bibinfo{person}{Minwei Feng},
  \bibinfo{person}{Cicero Nogueira~dos Santos}, \bibinfo{person}{Mo Yu},
  \bibinfo{person}{Bing Xiang}, \bibinfo{person}{Bowen Zhou}, {and}
  \bibinfo{person}{Yoshua Bengio}.} \bibinfo{year}{2017}\natexlab{}.
\newblock \showarticletitle{A structured self-attentive sentence embedding}.
\newblock \bibinfo{journal}{\emph{Proc. International Conference on Learning
  Representations (ICLR)}} (\bibinfo{year}{2017}).
\newblock


\bibitem[\protect\citeauthoryear{Lu, Xiong, Parikh, and Socher}{Lu
  et~al\mbox{.}}{2016}]%
        {Lu2016KnowingWT}
\bibfield{author}{\bibinfo{person}{Jiasen Lu}, \bibinfo{person}{Caiming Xiong},
  \bibinfo{person}{Devi Parikh}, {and} \bibinfo{person}{Richard Socher}.}
  \bibinfo{year}{2016}\natexlab{}.
\newblock \showarticletitle{Knowing When to Look: Adaptive Attention via a
  Visual Sentinel for Image Captioning}.
\newblock \bibinfo{journal}{\emph{Proc. the IEEE conference on computer vision
  and pattern recognition(CVPR)}} (\bibinfo{year}{2016}).
\newblock


\bibitem[\protect\citeauthoryear{Merity, Xiong, Bradbury, and Socher}{Merity
  et~al\mbox{.}}{2017}]%
        {Merity2016PointerSM}
\bibfield{author}{\bibinfo{person}{Stephen Merity}, \bibinfo{person}{Caiming
  Xiong}, \bibinfo{person}{James Bradbury}, {and} \bibinfo{person}{Richard
  Socher}.} \bibinfo{year}{2017}\natexlab{}.
\newblock \showarticletitle{Pointer Sentinel Mixture Models}.
\newblock \bibinfo{journal}{\emph{Proc. the International Conference on
  Learning Representations (ICLR)}} (\bibinfo{year}{2017}).
\newblock


\bibitem[\protect\citeauthoryear{Pan, Liang, Zhang, Yi, Yu, and Zheng}{Pan
  et~al\mbox{.}}{2018}]%
        {Pan2018HyperSTNetHF}
\bibfield{author}{\bibinfo{person}{Zheyi Pan}, \bibinfo{person}{Yuxuan Liang},
  \bibinfo{person}{Junbo Zhang}, \bibinfo{person}{Xiuwen Yi},
  \bibinfo{person}{Yingrui Yu}, {and} \bibinfo{person}{Yu Zheng}.}
  \bibinfo{year}{2018}\natexlab{}.
\newblock \showarticletitle{HyperST-Net: Hypernetworks for Spatio-Temporal
  Forecasting}.
\newblock \bibinfo{journal}{\emph{Proc. the Association for the Advancement of
  Artificial ntelligence (AAAI)}} (\bibinfo{year}{2018}).
\newblock


\bibitem[\protect\citeauthoryear{Radford, Wu, Child, Luan, Amodei, and
  Sutskever}{Radford et~al\mbox{.}}{2019}]%
        {radford2019gpt2}
\bibfield{author}{\bibinfo{person}{Alec Radford}, \bibinfo{person}{Jeff Wu},
  \bibinfo{person}{Rewon Child}, \bibinfo{person}{David Luan},
  \bibinfo{person}{Dario Amodei}, {and} \bibinfo{person}{Ilya Sutskever}.}
  \bibinfo{year}{2019}\natexlab{}.
\newblock \showarticletitle{Language Models are Unsupervised Multitask
  Learners}.
\newblock \bibinfo{journal}{\emph{OpenAI Blog}} (\bibinfo{year}{2019}).
\newblock


\bibitem[\protect\citeauthoryear{Seo, Kembhavi, Farhadi, and Hajishirzi}{Seo
  et~al\mbox{.}}{2016}]%
        {seo2016bidirectional}
\bibfield{author}{\bibinfo{person}{Minjoon Seo}, \bibinfo{person}{Aniruddha
  Kembhavi}, \bibinfo{person}{Ali Farhadi}, {and} \bibinfo{person}{Hannaneh
  Hajishirzi}.} \bibinfo{year}{2016}\natexlab{}.
\newblock \showarticletitle{Bidirectional attention flow for machine
  comprehension}.
\newblock \bibinfo{journal}{\emph{Proc. the International Conference on
  Learning Representations (ICLR)}} (\bibinfo{year}{2016}).
\newblock


\bibitem[\protect\citeauthoryear{Srivastava, Hinton, Krizhevsky, Sutskever, and
  Salakhutdinov}{Srivastava et~al\mbox{.}}{2014}]%
        {sariva2014dropout}
\bibfield{author}{\bibinfo{person}{Nitish Srivastava},
  \bibinfo{person}{Geoffrey Hinton}, \bibinfo{person}{Alex Krizhevsky},
  \bibinfo{person}{Ilya Sutskever}, {and} \bibinfo{person}{Ruslan
  Salakhutdinov}.} \bibinfo{year}{2014}\natexlab{}.
\newblock \showarticletitle{Dropout: A Simple Way to Prevent Neural Networks
  from Overfitting}.
\newblock \bibinfo{journal}{\emph{Journal of Machine Learning Research (JMLR)}}
  (\bibinfo{year}{2014}).
\newblock


\bibitem[\protect\citeauthoryear{Tang, Qu, Wang, Zhang, Yan, and Mei}{Tang
  et~al\mbox{.}}{2015}]%
        {Tang2015LINE}
\bibfield{author}{\bibinfo{person}{Jian Tang}, \bibinfo{person}{Meng Qu},
  \bibinfo{person}{Mingzhe Wang}, \bibinfo{person}{Ming Zhang},
  \bibinfo{person}{Jun Yan}, {and} \bibinfo{person}{Qiaozhu Mei}.}
  \bibinfo{year}{2015}\natexlab{}.
\newblock \showarticletitle{Line: Large-scale information network embedding}.
  In \bibinfo{booktitle}{\emph{Proc. the International Conference on World Wide
  Web (WWW)}}. \bibinfo{pages}{1067--1077}.
\newblock


\bibitem[\protect\citeauthoryear{Truong, Oudre, and Vayatis}{Truong
  et~al\mbox{.}}{2018}]%
        {Truong18}
\bibfield{author}{\bibinfo{person}{Charles Truong}, \bibinfo{person}{Laurent
  Oudre}, {and} \bibinfo{person}{Nicolas Vayatis}.}
  \bibinfo{year}{2018}\natexlab{}.
\newblock \showarticletitle{ruptures: change point detection in Python}.
\newblock \bibinfo{journal}{\emph{arXiv preprint arXiv:1801.00826}}
  (\bibinfo{year}{2018}).
\newblock


\bibitem[\protect\citeauthoryear{van~den Oord, Dieleman, Zen, Simonyan,
  Vinyals, Graves, Kalchbrenner, Senior, and Kavukcuoglu}{van~den Oord
  et~al\mbox{.}}{[n.d.]}]%
        {vanwavenet}
\bibfield{author}{\bibinfo{person}{A{\"a}ron van~den Oord},
  \bibinfo{person}{Sander Dieleman}, \bibinfo{person}{Heiga Zen},
  \bibinfo{person}{Karen Simonyan}, \bibinfo{person}{Oriol Vinyals},
  \bibinfo{person}{Alex Graves}, \bibinfo{person}{Nal Kalchbrenner},
  \bibinfo{person}{Andrew Senior}, {and} \bibinfo{person}{Koray Kavukcuoglu}.}
  \bibinfo{year}{[n.d.]}\natexlab{}.
\newblock \showarticletitle{WaveNet: A Generative Model for Raw Audio}. In
  \bibinfo{booktitle}{\emph{9th ISCA Speech Synthesis Workshop}}.
  \bibinfo{pages}{125--125}.
\newblock


\bibitem[\protect\citeauthoryear{Vaswani, Shazeer, Parmar, Uszkoreit, Jones,
  Gomez, Kaiser, and Polosukhin}{Vaswani et~al\mbox{.}}{2017}]%
        {Vas2017transformer}
\bibfield{author}{\bibinfo{person}{Ashish Vaswani}, \bibinfo{person}{Noam
  Shazeer}, \bibinfo{person}{Niki Parmar}, \bibinfo{person}{Jakob Uszkoreit},
  \bibinfo{person}{Llion Jones}, \bibinfo{person}{Aidan~N. Gomez},
  \bibinfo{person}{Lukasz Kaiser}, {and} \bibinfo{person}{Illia Polosukhin}.}
  \bibinfo{year}{2017}\natexlab{}.
\newblock \showarticletitle{Attention Is All You Need}. In
  \bibinfo{booktitle}{\emph{Proc. the Advances in Neural Information Processing
  Systems (NIPS)}}.
\newblock


\bibitem[\protect\citeauthoryear{Velickovic, Cucurull, Casanova, Romero,
  Li{\'o}, and Bengio}{Velickovic et~al\mbox{.}}{2018}]%
        {Veli2018gat}
\bibfield{author}{\bibinfo{person}{Petar Velickovic}, \bibinfo{person}{Guillem
  Cucurull}, \bibinfo{person}{Arantxa Casanova}, \bibinfo{person}{Adriana
  Romero}, \bibinfo{person}{Pietro Li{\'o}}, {and} \bibinfo{person}{Yoshua
  Bengio}.} \bibinfo{year}{2018}\natexlab{}.
\newblock \showarticletitle{Graph Attention Networks}.
\newblock \bibinfo{journal}{\emph{Proc. International Conference on Learning
  Representations (ICLR)}}  \bibinfo{volume}{abs/1710.10903}
  (\bibinfo{year}{2018}).
\newblock


\bibitem[\protect\citeauthoryear{Vlahogianni, Karlaftis, and
  Golias}{Vlahogianni et~al\mbox{.}}{2014}]%
        {Vlahogianni14}
\bibfield{author}{\bibinfo{person}{Eleni~I Vlahogianni},
  \bibinfo{person}{Matthew~G Karlaftis}, {and} \bibinfo{person}{John~C
  Golias}.} \bibinfo{year}{2014}\natexlab{}.
\newblock \showarticletitle{Short-term traffic forecasting: Where we are and
  where we’re going}.
\newblock \bibinfo{journal}{\emph{Transportation Research Part C: Emerging
  Technologies}} \bibinfo{volume}{43}, \bibinfo{number}{Part 1}
  (\bibinfo{year}{2014}), \bibinfo{pages}{3--19}.
\newblock


\bibitem[\protect\citeauthoryear{Wu, Pan, Long, Jiang, and Zhang}{Wu
  et~al\mbox{.}}{2019}]%
        {Wu2019GraphWF}
\bibfield{author}{\bibinfo{person}{Zonghan Wu}, \bibinfo{person}{Shirui Pan},
  \bibinfo{person}{Guodong Long}, \bibinfo{person}{Jing Jiang}, {and}
  \bibinfo{person}{Chengqi Zhang}.} \bibinfo{year}{2019}\natexlab{}.
\newblock \showarticletitle{Graph WaveNet for Deep Spatial-Temporal Graph
  Modeling}. In \bibinfo{booktitle}{\emph{Proc. the International Joint
  Conference on Artificial Intelligence (IJCAI)}}.
\newblock


\bibitem[\protect\citeauthoryear{Xu, Ba, Kiros, Cho, Courville, Salakhutdinov,
  Zemel, and Bengio}{Xu et~al\mbox{.}}{2015}]%
        {Xu2015ShowAA}
\bibfield{author}{\bibinfo{person}{Kelvin Xu}, \bibinfo{person}{Jimmy Ba},
  \bibinfo{person}{Ryan Kiros}, \bibinfo{person}{Kyunghyun Cho},
  \bibinfo{person}{Aaron~C. Courville}, \bibinfo{person}{Ruslan~R.
  Salakhutdinov}, \bibinfo{person}{Richard~S. Zemel}, {and}
  \bibinfo{person}{Yoshua Bengio}.} \bibinfo{year}{2015}\natexlab{}.
\newblock \showarticletitle{Show, Attend and Tell: Neural Image Caption
  Generation with Visual Attention}. In \bibinfo{booktitle}{\emph{Proc. the
  International Conference on Machine Learning (ICML)}}.
\newblock


\bibitem[\protect\citeauthoryear{Yu, Yin, and Zhu}{Yu et~al\mbox{.}}{2018}]%
        {yu2018spatio}
\bibfield{author}{\bibinfo{person}{Bing Yu}, \bibinfo{person}{Haoteng Yin},
  {and} \bibinfo{person}{Zhanxing Zhu}.} \bibinfo{year}{2018}\natexlab{}.
\newblock \showarticletitle{Spatio-temporal Graph Convolutional Networks: A
  Deep Learning Framework for Traffic Forecasting}. In
  \bibinfo{booktitle}{\emph{Proc. the International Joint Conference on
  Artificial Intelligence (IJCAI)}}.
\newblock


\bibitem[\protect\citeauthoryear{Zhang, Shi, Xie, Ma, King, and Yeung}{Zhang
  et~al\mbox{.}}{2018}]%
        {Zhang2018gaan}
\bibfield{author}{\bibinfo{person}{Jiani Zhang}, \bibinfo{person}{Xingjian
  Shi}, \bibinfo{person}{Junyuan Xie}, \bibinfo{person}{Hao Ma},
  \bibinfo{person}{Irwin King}, {and} \bibinfo{person}{Dit-Yan Yeung}.}
  \bibinfo{year}{2018}\natexlab{}.
\newblock \showarticletitle{GaAN: Gated Attention Networks for Learning on
  Large and Spatiotemporal Graphs}. In \bibinfo{booktitle}{\emph{Proc. the
  conference on uncertainty in artificial intelligence (UAI)}}.
\newblock


\bibitem[\protect\citeauthoryear{Zhao, Chen, Wu, Chen, and Liu}{Zhao
  et~al\mbox{.}}{2017}]%
        {Zhao17}
\bibfield{author}{\bibinfo{person}{Zheng Zhao}, \bibinfo{person}{Weihai Chen},
  \bibinfo{person}{Xingming Wu}, \bibinfo{person}{Peter~CY Chen}, {and}
  \bibinfo{person}{Jingmeng Liu}.} \bibinfo{year}{2017}\natexlab{}.
\newblock \showarticletitle{LSTM network: a deep learning approach for
  short-term traffic forecast}.
\newblock \bibinfo{journal}{\emph{IET Intelligent Transport Systems}}
  \bibinfo{volume}{11}, \bibinfo{number}{2} (\bibinfo{year}{2017}),
  \bibinfo{pages}{68--75}.
\newblock


\bibitem[\protect\citeauthoryear{Zheng, Fan, Wang, and Qi}{Zheng
  et~al\mbox{.}}{2019}]%
        {zheng2019gman}
\bibfield{author}{\bibinfo{person}{Chuanpan Zheng}, \bibinfo{person}{Xiaoliang
  Fan}, \bibinfo{person}{Cheng Wang}, {and} \bibinfo{person}{Jianzhong Qi}.}
  \bibinfo{year}{2019}\natexlab{}.
\newblock \showarticletitle{Gman: A graph multi-attention network for traffic
  prediction}.
\newblock \bibinfo{journal}{\emph{Proc. the Association for the Advancement of
  Artificial ntelligence (AAAI)}} (\bibinfo{year}{2019}).
\newblock


\bibitem[\protect\citeauthoryear{Zivot and Wang}{Zivot and Wang}{2006}]%
        {zivot2006vector}
\bibfield{author}{\bibinfo{person}{Eric Zivot} {and} \bibinfo{person}{Jiahui
  Wang}.} \bibinfo{year}{2006}\natexlab{}.
\newblock \showarticletitle{Vector autoregressive models for multivariate time
  series}.
\newblock \bibinfo{journal}{\emph{Modeling Financial Time Series with
  S-Plus{\textregistered}}} (\bibinfo{year}{2006}).
\newblock


\end{thebibliography}

\end{document}